\documentclass[a4paper]{article}
\usepackage[left=2.5cm,right=2.5cm,top=2cm,bottom=2cm]{geometry}
\usepackage{url} 
\usepackage{natbib}
\usepackage{caption}
\usepackage{subcaption}
\usepackage{amsmath}
\usepackage{amsfonts}
\usepackage{graphicx}
\usepackage{wrapfig}
\usepackage{lscape}
\usepackage{rotating}
\usepackage{multirow}
\usepackage{float}
\usepackage{authblk}
\usepackage{xcolor}
\usepackage{siunitx}

\begin{document}

\title{\hrule

\vspace{10pt}

\Large \textbf{An unsupervised learning approach for predicting wind farm power and downstream wakes using weather patterns}

\vspace{10pt}

\hrule}

\author[1,2]{Mariana C. A. Clare}
\author[2]{Simon C. Warder}
\author[3]{Robert Neal}
\author[4]{B. Bhaskaran}
\author[2]{Matthew D. Piggott}

\affil[1]{ECMWF, Bonn, Germany}
\affil[2]{Department of Earth Science and Engineering, Imperial College London, London, UK}
\affil[3]{Department of Weather Science, UK Meteorological Office, Exeter, UK}
\affil[4]{Shell Research Ltd, UK}

\date{}
\maketitle

\begin{abstract}
Wind energy resource assessment typically requires the use of numerical models, but such models are too computationally intensive to consider multi-year or multi-decade timescales. Increasingly, unsupervised machine learning techniques are used to identify a small number of representative weather patterns that can help simulate long-term behaviour. Here we develop a novel wind energy workflow that for the first time combines weather patterns derived from unsupervised clustering techniques with numerical weather prediction models (in our case WRF) to obtain efficient and accurate long-term predictions of power and downstream wakes from the entire wind farm. We use ERA5 reanalysis data and not only cluster on low altitude pressure, but also, for the first time, construct clusters using wind velocity, a more relevant variable for wind energy resource assessment. We also compare the use of large-scale (Europe-wide) and local-scale domains for the clustering. A WRF simulation is run at each of the cluster centres and the results aggregated into a long-term prediction using a novel post-processing technique. Through the application of our workflow to two case study regions, we show that, for a given variable and domain size combination, our long term predictions achieve excellent agreement with those from a year of WRF simulations, whilst requiring less than 2\% of the computational time. The most accurate results are obtained when clustering on wind velocity. Moreover, clustering over the Europe-wide domain is sufficient for predicting wind farm power output, but downstream wake predictions benefit from the use of smaller domains. Finally, we show that these downstream wakes themselves affect the local weather patterns. 

Our approach facilitates multi-year predictions of power output and downstream farm wakes, by providing a fast, accurate and flexible methodology that is applicable to any global region. Moreover, these accurate long-term predictions of downstream wakes provide the first tool to help mitigate the effects of wind energy loss downstream of wind farms, since they can be used to determine optimum wind farm locations.
\end{abstract}


\maketitle


\section{Introduction}\label{intro}
As the world attempts to transition to renewable energy over the coming decades, the global cumulative installed wind energy capacity will increase. A recent report by the International Renewable Energy Agency \citep{IRENA} suggests that in order to meet the Paris Climate Goals, offshore wind energy capacity needs to increase forty-fold by 2050 in order to meet both demand and clean energy goals. Hence, governments and energy companies globally are increasing investment in large-scale offshore wind farms. However, the build-out of wind farms must be carefully planned, not least because of the wind energy loss phenomenon sometimes referred to as `wind theft' where entire wind farms cause large scale wind velocity deficits downstream \citep{platis2018first,canadillas2022offshore}, notably decreasing the efficiency of wind farms located downstream \citep{Rajewski2014, Zhou2012, schneemann2020cluster,canadillas2020offshore,Pryor2020, Pryor2022,cuevas2022accuracy}. This effect is caused by wind farms extracting energy from the atmosphere and the physical consequence is referred to hereafter as the downstream wake. This wake has the potential to significantly impact the optimum wind farm location and spatial arrangement -- \cite{Pryor2021} predict that downstream wakes from future farms at the current proposed sites in the US could reduce overall national offshore wind production in the US by one third of the expected value. This in turn leads to large economic losses, and stresses the need for the coordination of future wind energy development (including across national borders) \citep{Lundquist2018,schneemann2020cluster,energiewende2020making,fischereit2022review}, especially given increased farm densities in the future \citep{akhtar2021accelerating}.

Numerical models such as the Weather Research and Forecasting (WRF) model \citep{powers2017weather} can be used to simulate the downstream wakes caused by wind farms. However, these models struggle to provide a holistic picture because they are too computationally expensive to be run on multi-decadal or even multi-year scales. One approach to deal with this issue is to identify a small subset of simulation periods which can be considered representative of the overall wind climatology. For example, \citet{Pryor2021} and \citet{Pryor2022} identify `dominant modes' (\textit{i.e.} frequent combinations of wind speed and direction) and perform a series of five-day simulations for each. Aggregated predictions are then obtained through the use of weighted averages based on the frequency of the occurrence of each `mode'. In an alternative approach, \citet{kumar2021large} and \citet{cuevas2022accuracy} identify simulation periods of two or more weeks which, when aggregated, capture similar probability distributions of wind speed and direction to the overall wind climatology over a longer period. However, both approaches discussed suffer due to the difficulties in identifying representative weather patterns, particularly when these representative patterns are based purely on temporal wind speed and direction frequencies and do not take into account spatial variability.

Recently, there has been growing research into identifying representative weather patterns more rigorously using unsupervised machine learning clustering techniques, which account for both spatial and temporal variations. For example, \cite{ferranti2015flow,neal2016flexible} use patterns identified in this way to improve the use of weather information in downstream applications. Patterns identified using machine learning methods have also been used to predict extreme precipitation events \citep{howard2022weather, camus2022daily}, reconstruct wind speeds on specific days \citep{cortesi2019characterization, torralba2021challenges}, and for long-term wind power forecasting either using the average wind speeds from the clustered weather patterns with simple power curves \citep{cheneka2020impact} or comparing clustered weather patterns with known wind power production to find trends between the pattern and the power output \citep{garrido2020impact,cheneka2021associating}. 

So far, no research has combined rigorous clustering techniques with an accurate numerical model such as WRF to determine power estimates, or assess the issue of power losses due to wakes from upstream farms. Moreover, all clustering for wind energy production has focused on clustering based upon low-altitude pressure fields \citep[e.g.][]{neal2016flexible,garrido2020impact}, rather than directly on variables which are known to be of primary importance for wind power prediction such as wind speed and direction. There is therefore great scope to improve the accuracy of wind power predictions using clustering, and to use these techniques to predict the impact of downstream farm wakes. 

In this work, we develop a novel wind energy workflow that shows for the first time how complex numerical weather prediction models can be successfully combined with unsupervised clustering techniques to efficiently make accurate long-term predictions of wind farm power and downstream wakes. The first step of this workflow is to identify weather patterns by using the unsupervised k-means clustering technique applied to ERA5 reanalysis data, thus accounting for both spatial and temporal variations. A WRF simulation is then run using the average weather conditions in each cluster (\textit{i.e.} the weather conditions of the cluster centre) to determine the typical cluster power output and downstream wind farm wake. In contrast to existing literature \citep[see for example][]{garrido2020impact}, we do not simply derive clusters from low-altitude pressure but also derive clusters from \SI{100}{\metre} wind velocity, which is an assumed average hub height for wind turbines and is thus likely to be of more primary importance for wind energy prediction. Furthermore, we cluster the latter over spatial domains of two different geographical sizes in order to understand whether wind energy related predictions are improved by considering more local weather patterns. This analysis allows us to conclude which variable and which domain size are most appropriate for determining weather patterns for offshore wind energy production. Furthermore, once the WRF simulations are run, we also apply a novel post-processing approach to the cluster simulations to improve long-term predictions of both wind power production and downstream wakes. Finally, we also consider whether the downstream wakes caused by the wind farms are sufficient to change the weather patterns themselves. This is of great interest for long-term planning and build-out of wind farms, and any changes highlight the need for a flexible and efficient approach such as ours, which are sufficiently computationally efficient to manage evolving scenarios.

The key novelty of our approach is that it facilitates multi-year and/or multi-decadal predictions of the power and downstream wakes generated by a given offshore wind farm, without having to run a simulation for this entire length of time. Whilst other works have performed small-scale studies of downstream wind farm wakes, this is the first work which provides a tool that can help mitigate the effects of these wakes through providing accurate and fast long-term predictions that result in more accurate knowledge of optimum wind farm locations. Crucially, in this work, we evaluate the accuracy of the predictions obtained by our workflow by benchmarking them against an entire year of WRF simulations. We choose to do this at two different locations to show that the quality of our results is not location dependent. The first is to the west of Denmark and the second to the east of the Shetland Islands; both of these areas are known to be well suited to wind farm development \citep{newcombe2021orion,johansen2021blowing}. However, we emphasise here that the aim of this work is to create a tool which can be used for long-term predictions globally and our approach has therefore been specifically designed to be generalisable and readily applicable to any offshore region of interest in the world. 

The remainder of this work is structured as follows: in Section \ref{sec:methodology} we briefly outline the methodology used including the data, WRF model and the k-means clustering technique; in Section \ref{sec:clustering}, we use k-means clustering to determine weather patterns for different domains and different weather variables; in Section \ref{sec:patterns}, we use these weather patterns to predict wind farm power output and downstream wakes and conclude this work in Section \ref{sec:conclusion}.

\section{Data and Methods}\label{sec:methodology}
\begin{figure}[!t]
    \centering
    \includegraphics[width = 0.4\textwidth]{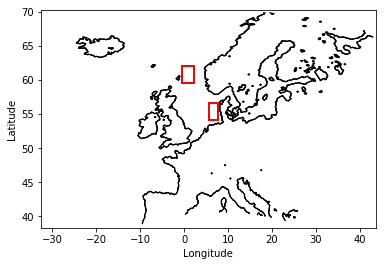}
    \caption{Extent of the larger European domain considered here, with the two smaller North Sea domains also shown and outlined in red.}
    \label{fig:domain_location}
\end{figure}

In this work, we develop a workflow to make long-term predictions of wind farm power and downstream wakes. As a first step, we cluster ERA5 reanalysis data with spatial resolution $0.25^{\circ}$ \citep{hersbach2020era5} at 1200 UTC daily over a 10-year period from 2000--2009 to determine weather patterns. This is a similar time period to that used to find clusters for wind energy in \cite{cheneka2020impact} and \cite{cheneka2021associating}, but further research could look at increasing the time period to improve climatological representativeness of the weather patterns. We emphasise, however, that the purpose of this work is not to find weather patterns that are sensible synoptically but to find ones that result in good estimates for power and downstream wakes. As mentioned in Section \ref{intro}, in the existing literature \citep[for example][]{neal2016flexible} weather pattern clustering is based on low-altitude pressure data. We hypothesise, however, that this is not the best variable for wind energy production and that instead \SI{100}{\metre} wind velocity (as a typical hub height for a wind turbine) would provide clusters that are better suited to our wind energy application. We therefore here consider clustering on both \SI{100}{\metre} wind velocity \citep{hersbach2018era5} and sea-level pressure \citep{hersbach2018era5_single} reanalysis data over a Europe-wide domain. Furthermore, we also cluster wind velocity on two smaller domains in the North Sea shown in Figure \ref{fig:domain_location} (one west of Denmark and the other east of the Shetland Islands) to understand if there are advantages from clustering on more local wind-farm scales. The difference between the weather patterns at the two smaller locations is discussed in detail in Section \ref{sec:clustering}.

For the clustering algorithm itself, we use k-means clustering \citep{likas2003global}. This aims to partition $n$ observations into $k$ clusters, by using an iterative process to minimise the least squared Euclidean distance between each cluster centre and the points assigned to that cluster. Here, we implement this algorithm using the scikit-learn python package \citep{scikit-learn}. K-means clustering is an unsupervised learning approach (\textit{i.e.}, the data is unlabelled) and thus the user must choose the number of clusters $k$ into which the data should be clustered. This is typically done by using the `elbow method' \citep{bholowalia2014ebk}, where some heuristic of cluster accuracy is considered as a function of the number of clusters, and the number of clusters is then chosen to be the point at which there is an `elbow' in this relationship. An example of these elbow curves is shown in Figures \ref{fig:cluster_elbow} in Section \ref{sec:clustering}, with discussion of the cluster accuracy heuristics used in this work also contained in that section.

\begin{figure}[!t]
    \centering
    \includegraphics[width = 0.45\textwidth]{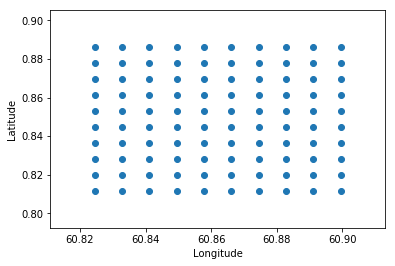}
    \caption{Hypothetical wind farm array layout used in this work. There are 100 wind turbines with a maximum power rating of 5000kW. Note the latitudes and longitudes shown are for the Shetland Islands domain, but the array layout is exactly the same for the Denmark domain.}
    \label{fig:wind_farm}
\end{figure}

Once we have obtained the clusters, the second step of our workflow is to use WRF (Weather Research and Forecasting Model) \citep{skamarock2019description} to simulate both the power production of, and downstream wake from, a hypothetical wind farm, for weather conditions representative of each cluster. WRF is a fully compressible non-hydrostatic numerical weather prediction (NWP) model, which is widely used for both atmospheric research and operational weather forecasting \citep{powers2017weather}. For our simulations, the initial conditions for the WRF model are the weather conditions corresponding to the datapoint which is closest to each cluster centre. Note that the closest datapoint $Y_{i}$ is defined as the one which minimises $\lvert\lvert Y_{i} - \bar{Y} \rvert \rvert^{2}$, where $\bar{Y}$ is the cluster centre. We also conduct the WRF simulations over the whole Europe-wide domain, even if the clusters themselves have been generated by clustering on the smaller domain. Finally we place the hypothetical wind farm comprising of 100 5000kW turbines at the centre of either the small Denmark domain or the small Shetland Islands domain, with the array spacing shown in Figure \ref{fig:wind_farm}. WRF is run once for each cluster and for reasons of computational cost, each WRF run within this work spans a period of three hours, which is more than sufficient for WRF to spin up and simulate the resulting downstream wind farm wake.

 The WRF model has the valuable feature that it allows for nested domains with progressively finer spatial resolutions, a process known as dynamical downscaling. This means that it is possible to simulate local weather around the wind farm, as well as larger scale regional weather processes. In this work, we use a two domain approach shown in Figure \ref{fig:domain_location} -- the larger one corresponds to the Europe-wide domain (for which we use a resolution of 16.6km) and the smaller is a nested domain spanning a localised region containing the wind-farms being considered (for which we use a resolution of 1.8km). However, even with these nested domains, modelling both large-scale weather patterns and local scale flow interactions within the wind farms themselves is both difficult and computationally expensive. Hence WRF contains a parametrisation scheme to simulate the presence of wind farms, known as the Fitch scheme \citep{fitch2012local}. The Fitch parametrisation explicitly excludes the accurate representation of near-wake dynamics and instead is designed to simulate the relatively-far-wake dynamics produced from all of the wind turbines that are located within a WRF grid cell \citep{pryor2019wind}. This is done by imposing a drag force on all grid cells affected by the wind farm, which represents the kinetic energy extracted by the wind turbine from the flow. The drag is calculated as
\begin{equation}
    F_{\text{drag}} = \frac{1}{2}C_{T}\rho A \mathbf{V} \lvert \mathbf{V}\rvert,
\end{equation}
where $C_{T}$ is a turbine thrust coefficient dependent on the wind speed and turbine type, $\mathbf{V}$ is the wind velocity and $A$ is the cross-sectional rotor area. The power produced by the wind turbines in each grid cell is then estimated from a turbine specific power curve using the wind speed profile averaged across the grid-cell \citep{Pryor2020}. We refer the reader to \cite{fitch2012local} for more details.

\subsection{Combining WRF results for clusters to obtain a long-term prediction}
An important attribute of k-means clustering is that the algorithm assigns each gridded field to a cluster in an unsupervised approach. Note that, in order to be consistent with machine learning terminology, hereafter we refer to the gridded fields as datapoints. A simple method to obtain long-term predictions of wind farm power and downstream wakes from the WRF results at the cluster centres is to add them together in a weighted sum given by
\begin{equation}\label{eq:simple}
    Y_{\text{long-term}} = \sum_{i = 1}^{C_\text{tot}} N_{i} Y_{i},
\end{equation}
where $Y$ is the two-dimensional field of interest (here either the difference in the wind field due to the presence of the wind farm, or the power generated by said wind farm), $Y_{i}$ is that field computed for cluster $i$, $C_\text{tot}$ represents the number of clusters and $N_{i}$ is number of datapoints in cluster $i$. In Section \ref{sec:patterns}, we show the long-term predictions obtained using this method.

A drawback of this simple approach that collapses down fully time-varying information to a small number (e.g. six) of clusters is that it will artificially inflate the downstream wake in the directions of the finite number of cluster centres, creating an almost discontinuous (in angle) wake field. In reality, the wake field should be relatively smooth due to the natural variability in wind direction. Furthermore, the reanalysis data (used for the clustering step) provides values for the initial wind fields at 1200UTC in the absence of wind farms, and this information can be used to improve on the simple cluster-derived prediction given by equation \ref{eq:simple}. Thus, we can consider a more complex approach, where for each datapoint we `correct' the cluster WRF prediction for power and downstream wake using the ratio between the initial wind direction and speed at 1200UTC for that datapoint and the cluster. Note that we use the initial wind fields for the ratio, so as to ensure consistent data sources -- both sides of the ratio are derived from the ERA5 data.

First let us consider how to modify the long-term power prediction. At the resolution we are considering, there is essentially no spatial variability in the power production across the wind farm and therefore we only need to modify the magnitude of the power. The long-term prediction for power, $P_{\text{long-term}}$ is thus
\begin{equation}
    P_{\text{long-term}} = \sum_{i = 1}^{C_\text{tot}} \sum_{j=1}^{N_{i}} \frac{\lvert \mathbf{u}_{j}\rvert}{\lvert \mathbf{u}_{i}\rvert}P_{i},
\end{equation}
where $P_{i}$ is the WRF power field prediction for cluster $i$, $\lvert \mathbf{u}_{j}\rvert$ is the initial speed at the centre of the future wind farm for datapoint $j$ and $\lvert \mathbf{u}_{i}\rvert$ the initial velocity at the centre of the future wind farm for cluster $i$. We emphasise that we obtain $\mathbf{u}_{i}$ and $\mathbf{u}_{j}$ from the ERA5 reanalysis data and thus there is no actual wind farm present in these velocity fields. Note further that wind turbines have a maximum power rating they can achieve and therefore we impose the maximum power rating of the wind farm as a limit on each individual weighted power prediction.

For the long-term prediction of a farm's downstream wake, the direction of the wind field is also critical and therefore we must rotate the wake field as well as modify its magnitude. Consider a datapoint $j$ in cluster $i$, with $\theta_{j}$ the initial wind angle for datapoint $j$ at the centre of the future wind farm and $\theta_{i}$ the initial wind angle for cluster $i$ at the centre of the future wind farm. To obtain a prediction for the wake field of datapoint $j$, we rotate the wake field from cluster $i$ by $\theta_{j} - \theta_{i}$ using a standard rotation matrix. Moreover to correct the wake field magnitude, we then multiply the rotated wake field by the ratio between the cluster $i$ and the datapoint $j$ of the initial speeds at the centre of future wind farm. The long-term downstream wake prediction, $W_{\text{long-term}}$ is thus
\begin{equation}\label{eq:complex wake}
    W_{\text{long-term}} = \sum_{i = 1}^{C_{\text{tot}}} \sum_{j=1}^{N_{i}} \frac{\lvert \mathbf{u}_{j}\rvert}{\lvert\mathbf{u}_{i}\rvert}f(W_{i}; \theta_{i}, \theta_{j}),
\end{equation}
where $f(\cdot)$ is the function which rotates the wake-field, and $W_{i}$ is the wake field prediction for cluster $i$. We emphasise again that $\theta_{i}$, $\theta_{j}$, $\mathbf{u}_{i}$ and $\mathbf{u}_{j}$ come from the ERA5 reanalysis data and therefore no extra WRF simulations are required.

In Section \ref{sec:patterns}, we compare the long-term predictions from using this more complex approach with those from using the simple approach. We emphasise here that this more complex approach does not require further WRF simulations and should be seen as a computationally cheap post-processing step.

\section{Clustering weather data using k-means clustering}\label{sec:clustering}
In this work, we use k-means clustering to cluster weather data. As this is an unsupervised approach, we must first determine the number of clusters, $k$, that the data should be split into. As discussed in Section \ref{sec:methodology}, we do this using the `elbow method' with three different cluster accuracy heuristics: the average distance between the cluster centre and each point assigned to that cluster, the average correlation between the points in each cluster (using the Pearson product-moment correlation coefficient), and the silhouette score defined as 
\begin{equation}
    \text{silhouette score} = \frac{(b-a)}{\max(a,b)},
\end{equation}
where $a$ is the mean intra-cluster distance (\textit{i.e.}, the mean distance between points within a cluster) and $b$ is the mean distance between a point and the nearest cluster that it is not a part of \citep{shahapure2020cluster}. Note that the best silhouette score is 1, the worst -1 and 0 indicates that the clusters overlap.

\begin{figure}[!t]
\begin{subfigure}{0.95\textwidth}
    \centering
    \includegraphics[height=0.2\textwidth]{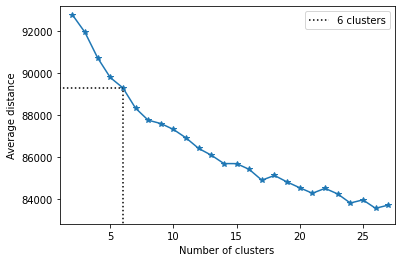}
    \hfill
    \includegraphics[height=0.2\textwidth]{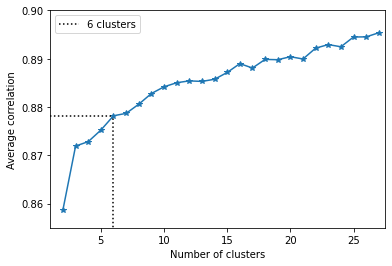}
    \hfill
    \includegraphics[height=0.2\textwidth]{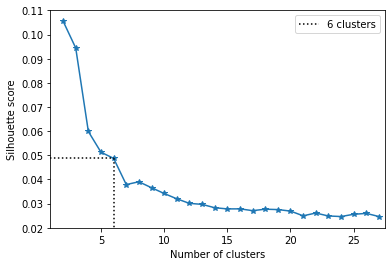}
    \caption{Clustering on wind velocity data from 2000--2009 on Europe-wide domain.}
    \end{subfigure}
\begin{subfigure}{0.95\textwidth}
    \centering
    \includegraphics[height=0.2\textwidth]{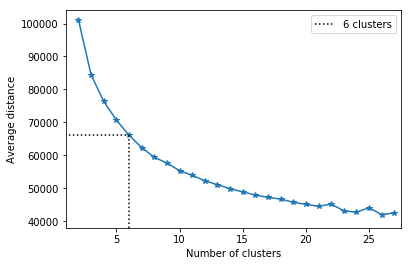}
    \hfill
    \includegraphics[height=0.2\textwidth]{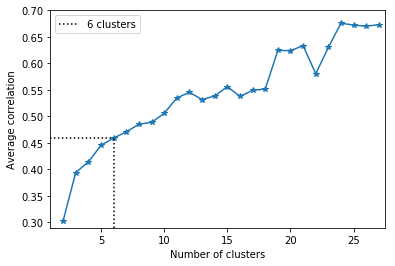}
    \hfill
    \includegraphics[height=0.2\textwidth]{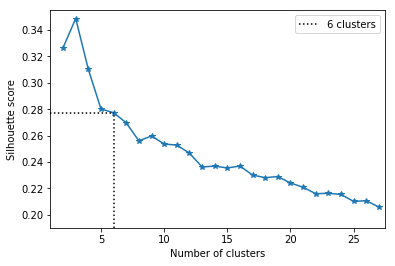}
    \caption{Clustering on wind velocity data from 2000--2009 on small Denmark domain.}
    \end{subfigure}
    \begin{subfigure}{0.95\textwidth}
    \centering
    \includegraphics[height=0.2\textwidth]{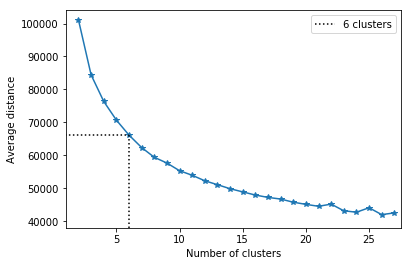}
    \hfill
    \includegraphics[height=0.2\textwidth]{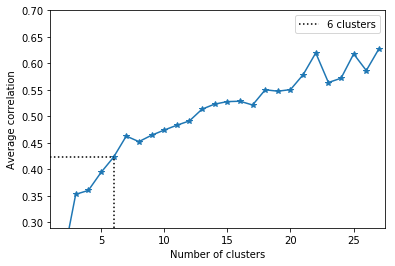}
    \hfill
    \includegraphics[height=0.2\textwidth]{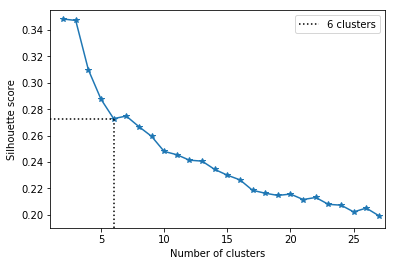}
    \caption{Clustering on wind velocity data from 2000--2009 on small Shetland Islands domain.}
    \end{subfigure}
    \begin{subfigure}{0.95\textwidth}
    \centering
    \includegraphics[height=0.2\textwidth]{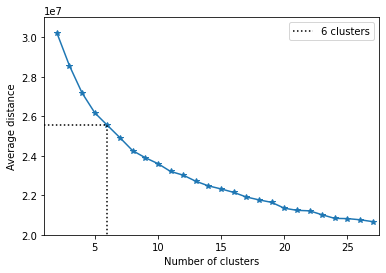}
    \hfill
    \includegraphics[height=0.2\textwidth]{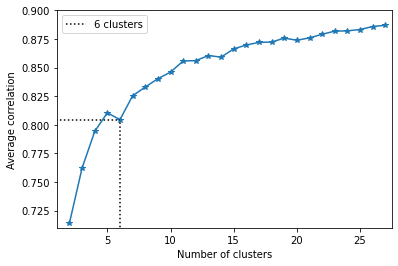}
    \hfill
    \includegraphics[height=0.2\textwidth]{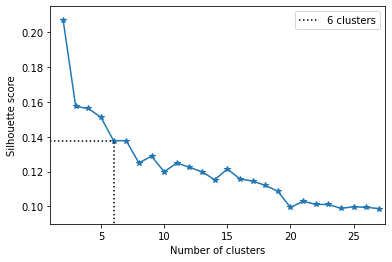}
    \caption{Clustering on sea-level pressure data from 2000--2009 on Europe-wide domain.}
   \end{subfigure}
    \caption{Elbow curves with different heuristics for range of different physical variables and domains. Left: Average distance between points in cluster and cluster centre. Centre: Correlation between points in each cluster. Right: Silhouette score (how well samples are clustered with other samples that are similar to each other).}\label{fig:cluster_elbow}
\end{figure}

As previously discussed in Section 2, we consider clustering on both sea-level pressure and 100m wind velocity separately, considering one domain size for the former and three domain sizes for the latter. However, we also needed to decide on an optimal number of clusters, and for this we interpret the results in the elbow curves (Figure \ref{fig:cluster_elbow}). For average distance a lower score is better, but for correlation and the silhouette score, a higher score is better. The average distance between points in each cluster and the cluster centres becomes smaller as the number of clusters increases, which is unsurprising as clusters are more likely to become similar as the number increases. This distance is also a reflection of the similarity in magnitude of features within each cluster, whereas correlation looks at the similarity in spatial mapping within each cluster.  Here, the average correlation between points in each cluster increases as the number of clusters increases. Again, this is to be expected as the sample size in each cluster reduces. Finally, the silhouette score trends closer to zero as the number of clusters increases. This is because it becomes increasingly likely that clusters will overlap. Despite the datasets being substantially different, six clusters generally seems to be a good choice for all three heuristics. This number of clusters is the same or very similar as that used in other studies looking at clusters for weather patterns such as \cite{neal2016flexible,garrido2020impact,falkena2020revisiting}, thus corroborating our choice.

\begin{figure}[!t]
\begin{subfigure}{0.95\textwidth}
    \centering
    \includegraphics[width=0.5\textwidth]{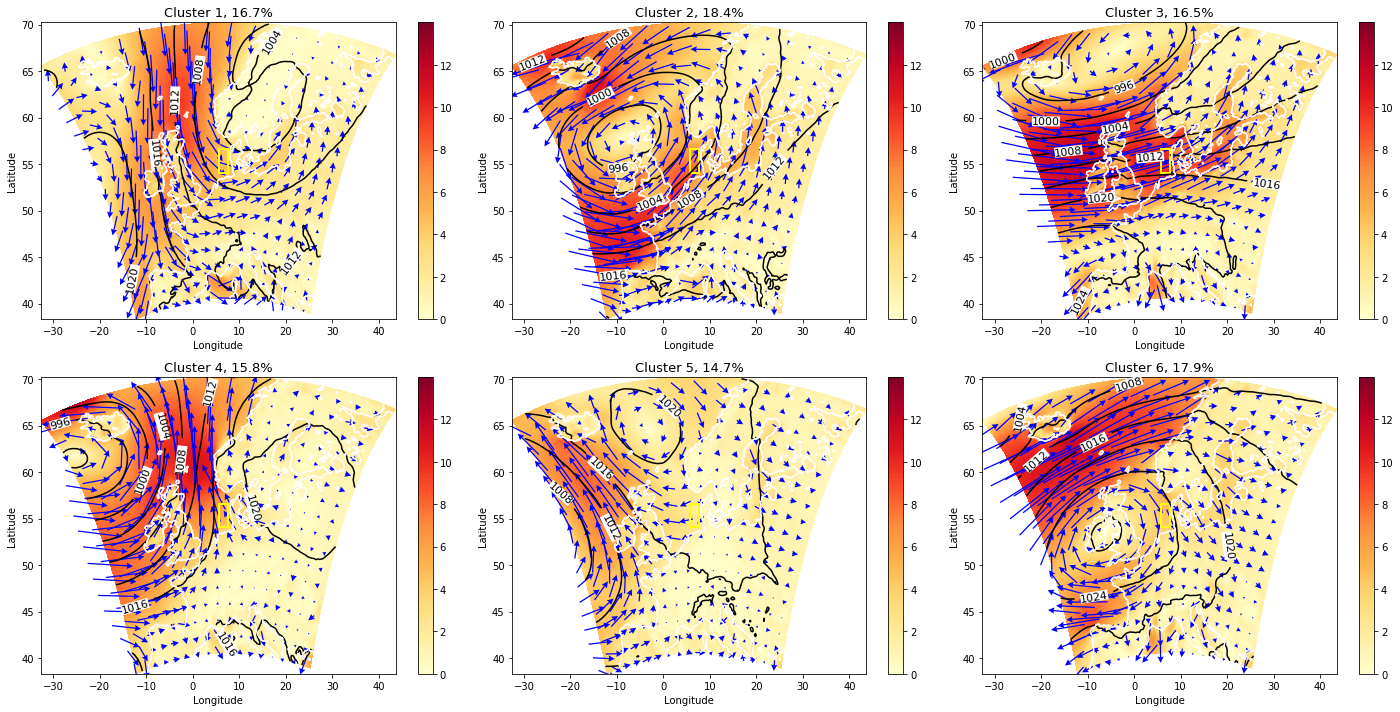}
    \caption{Clustering on wind velocity data in the Europe-wide domain.}
    \end{subfigure}
    \begin{subfigure}{0.95\textwidth}
    \centering
        \includegraphics[width=0.5\textwidth]{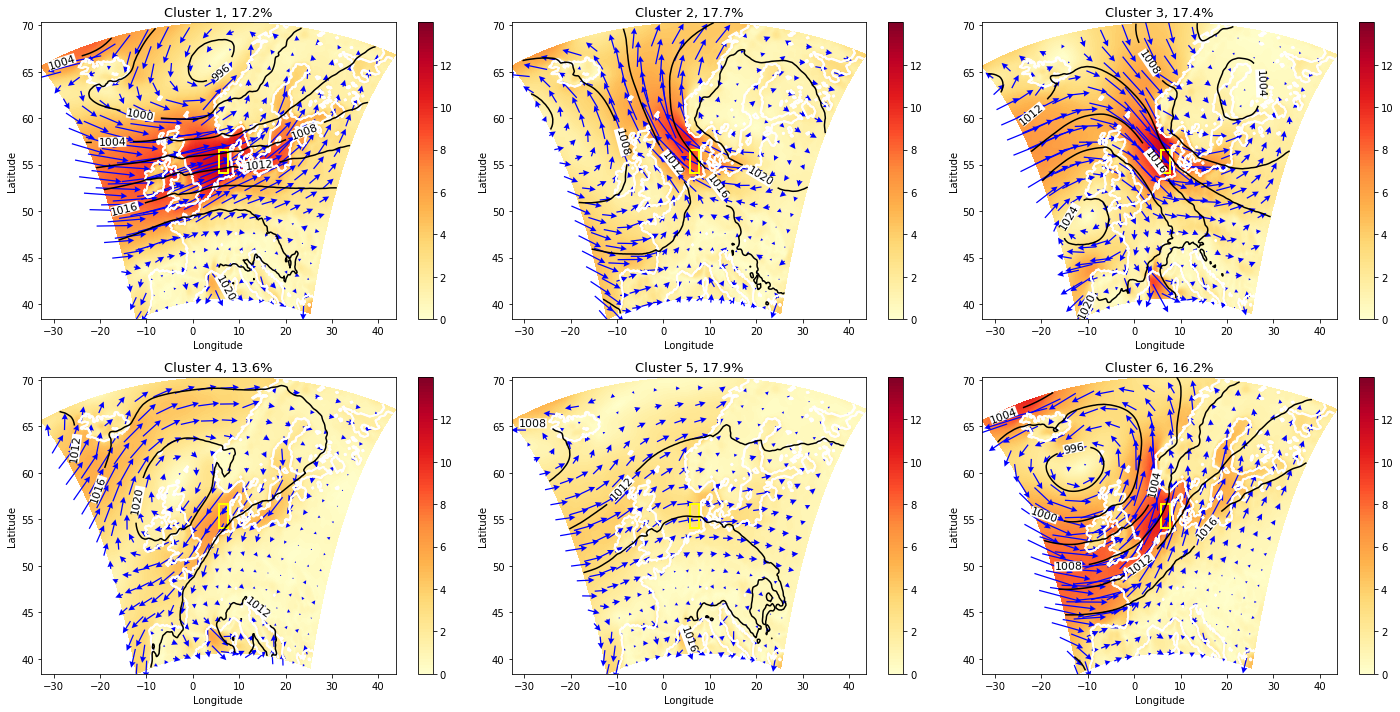}
    \caption{Average pressure and wind velocity for the patterns found by clustering on wind velocity data in the smaller Denmark domain.}
    \end{subfigure}
        \begin{subfigure}{0.95\textwidth}
        \centering
        \includegraphics[width=0.5\textwidth]{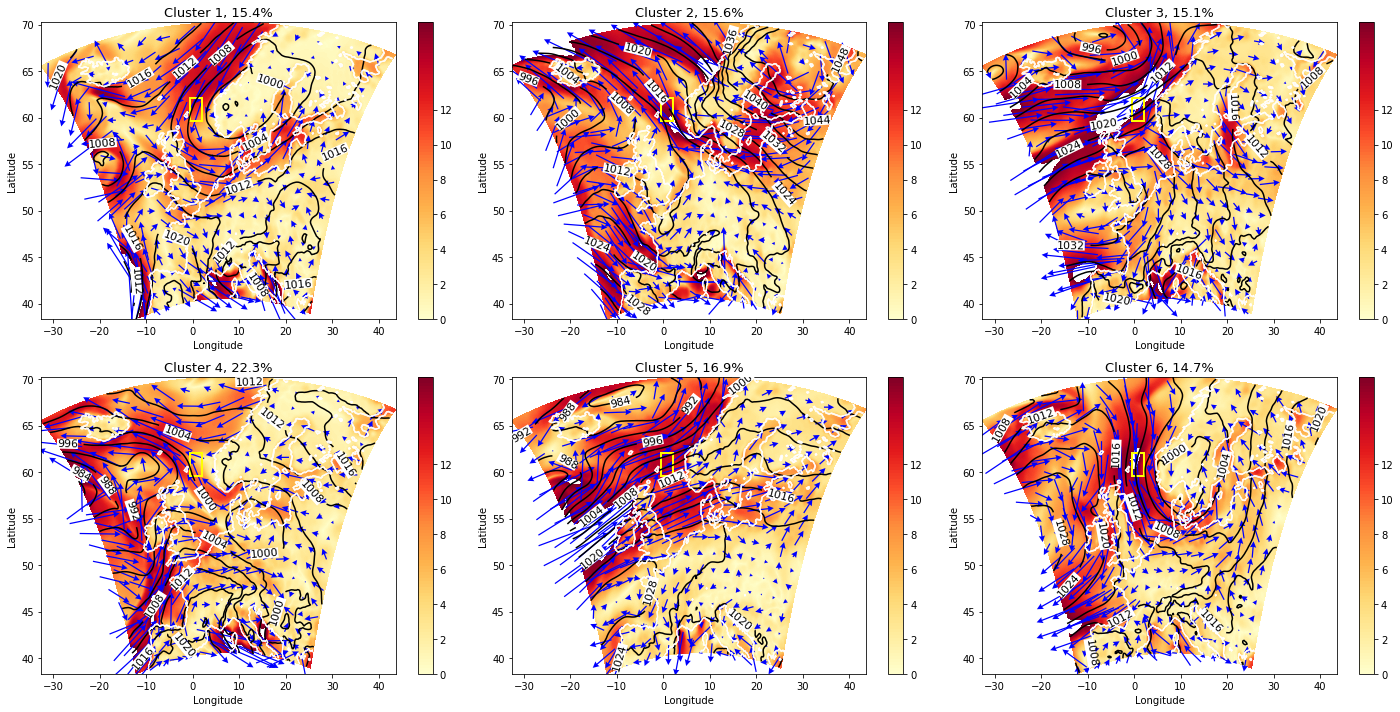}
    \caption{Average pressure and wind velocity for the patterns found by clustering on wind velocity data in the smaller Shetland Islands domain.}
    \end{subfigure}
    \begin{subfigure}{0.95\textwidth}
    \centering
    \includegraphics[width=0.5\textwidth]{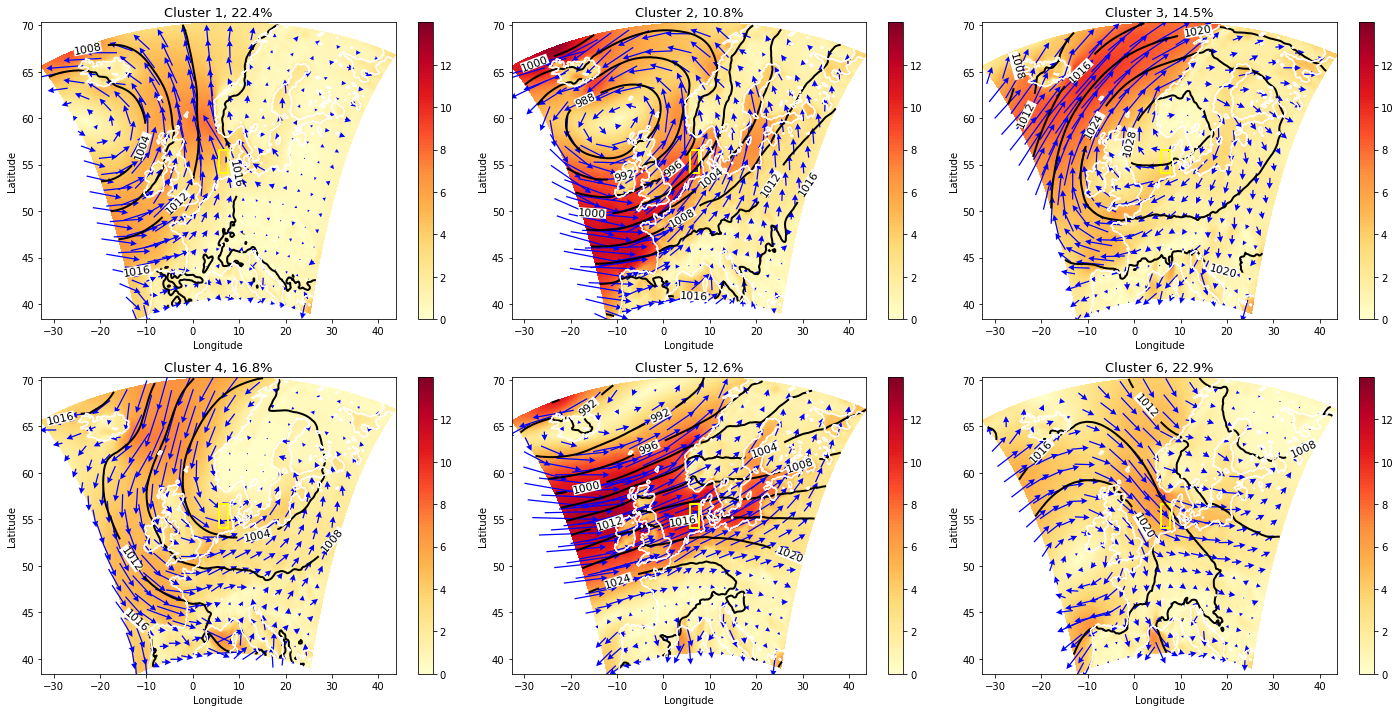}
    \caption{Clustering on sea-level pressure data in the Europe-wide domain.}
    \end{subfigure}
    \caption{Europe-wide domain clusters and averages using smaller domains. Filled contours represent the wind speed, black contours the pressure and the arrows the wind direction with size reflecting the wind speed magnitude.}\label{fig:cluster_d01}
\end{figure}

Recall that we use the closest real datapoint to the cluster centres for our WRF runs. We therefore know sea-level pressure and wind velocity for each domain at these datapoints and thus can compare their behaviour across the different cluster sets. Figure \ref{fig:cluster_d01} therefore not only shows the Europe-wide domain when clustering on wind velocity data (a) and clustering on sea-level pressure data (d) in this larger domain, but also what the larger domain looks like when clustering on wind velocity in the smaller Denmark domain (b) and in the Shetland Islands domain (c). It shows that the clustering on wind velocity on the Europe-wide domain and on the Denmark domain, leads to similar Europe-wide clusters. For example, Cluster 3 in (a) and Cluster 1 in (b) both show a mobile westerly flow over the UK with low pressure centred to the north of Scotland and high pressure centred over the Azores or central/southern Europe. Note that these two cluster centres are also very similar to Cluster 5 in (d) (found by clustering on pressure in the Europe-wide domain), which is a common North Atlantic-dominating weather pattern over Europe \citep{ferranti2015flow, neal2016flexible}. Other common weather patterns are also shown by these cluster sets, including northerlies (e.g. Cluster 1 (a), Cluster 3 (b) and Cluster 4 (d), south-westerlies (e.g. Cluster 4 (a), Cluster 6 (b) and Cluster 1 (d), and blocking high pressure patterns (e.g. Cluster 5 (a), Cluster 4 (b) and Cluster 3 (d). Generally, clustering on wind velocity in the Denmark domain leads to weaker mean speeds than clustering on wind velocity in the Europe-wide domain. In contrast, clustering on wind velocity in the Shetland Islands domain leads to much higher wind speeds and poorly defined weather patterns (i.e. it is difficult to match them up to the common patterns found in (a), (b) and (d). By far the clearest weather systems and pressure structures are found when clustering on pressure in the Europe-wide domain. The overall structure though is similar to that when clustering on wind velocity in the Europe-wide domain. However, the percentage of datapoints (days) which fall into each cluster differ. For example, when looking at the common mobile westerly type, Cluster 3 in (a) accounts for 16.5\% of days and Cluster 5 in (d) accounts for 12.6\% of days, which may result in some differences in the prediction of wind energy and downstream wakes. Ultimately, we emphasise that for this work, it is far more important to find weather patterns which can robustly estimate power output and downstream wind farm wake than to find patterns which are synoptically sensible.

\begin{figure}[!t]
\begin{subfigure}{0.95\textwidth}
    \centering
    \includegraphics[width = 0.8\textwidth]{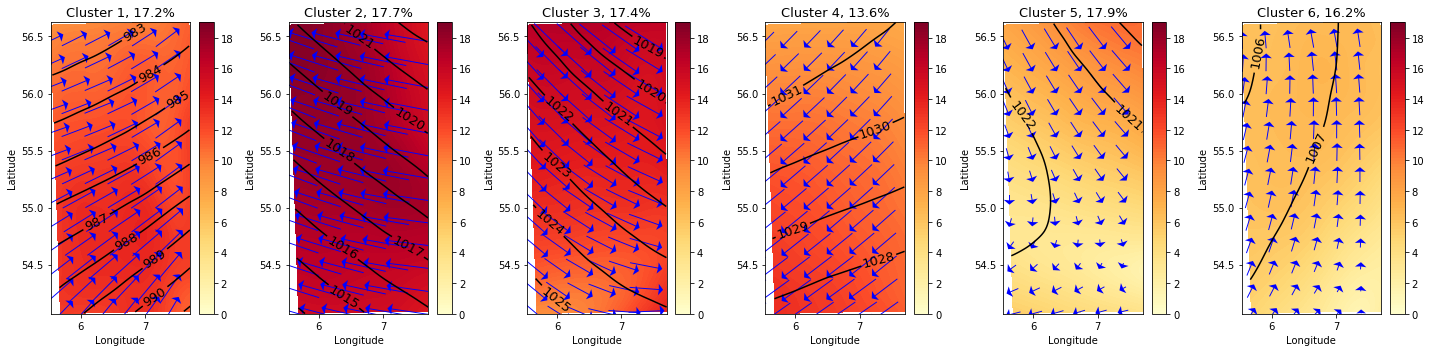}
    \caption{Clustering on wind velocity data in the smaller Denmark domain}
\end{subfigure}
\begin{subfigure}{0.95\textwidth}
    \centering
    \includegraphics[width = 0.8\textwidth]{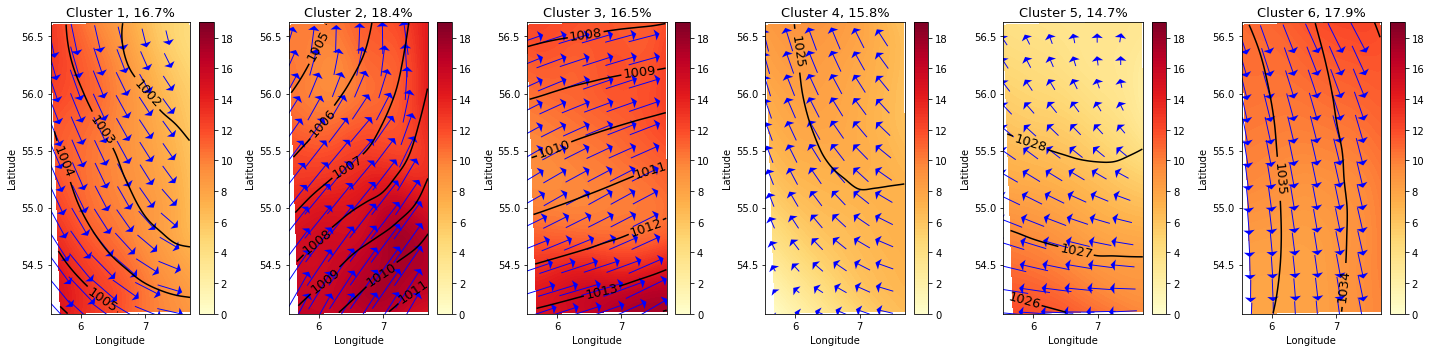}
    \caption{Average pressure and wind velocity for the patterns found by clustering on wind velocity data in the Europe-wide domain.}
\end{subfigure}
\begin{subfigure}{0.95\textwidth}
    \centering
    \includegraphics[width = 0.8\textwidth]{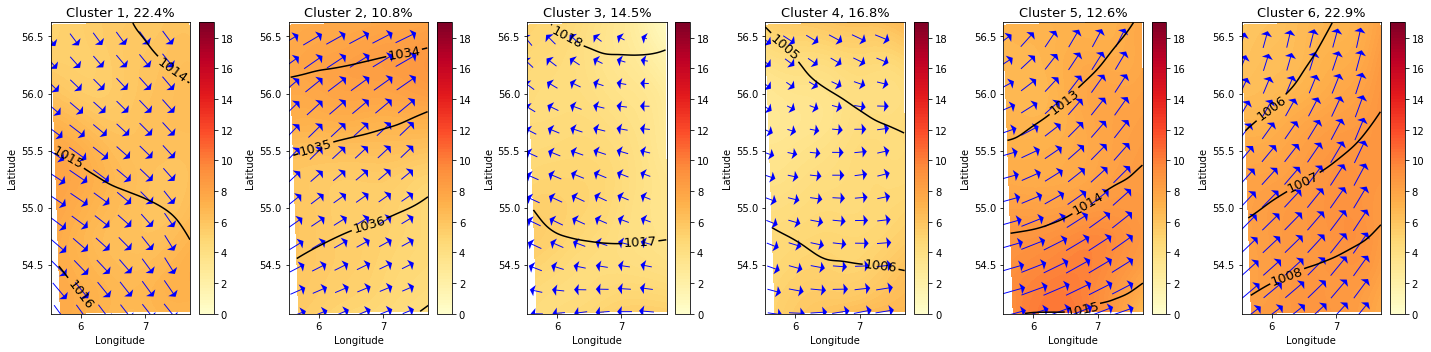}
    \caption{Average pressure and wind velocity for the patterns found by clustering on sea-level pressure date in the Europe-wide domain.}
\end{subfigure}
\caption{Denmark domain cluster and averages using Europe-wide domains. Filled contours represent the wind speed, black contours the pressure and the arrows the wind direction with size reflecting the wind speed magnitude.}\label{fig:cluster_d02_den}
\end{figure}

In Figure \ref{fig:cluster_d02_den}, the same analysis is shown but for the small Denmark domain with (a) showing the clusters when clustering on wind velocity data in that domain, and (b) the clusters when clustering on wind velocity and (c) on pressure in the Europe-wide domain. Unlike with the larger domain, each cluster set is very different. In particular, the clusters found from clustering on pressure in the larger domain have a much lower wind speed than those found by clustering on wind velocity data in either domain. Whilst clustering on wind velocity in the larger (a) and smaller domain (b) results in clusters of similar magnitude, the wind patterns are very different. This is a good indication that the predictions for wind energy production and downstream wakes will be different in the Denmark domain, depending on the size of domain and variable on which the clustering is conducted.

\begin{figure}[!t]
\begin{subfigure}{0.95\textwidth}
    \centering
    \includegraphics[width = 0.8\textwidth]{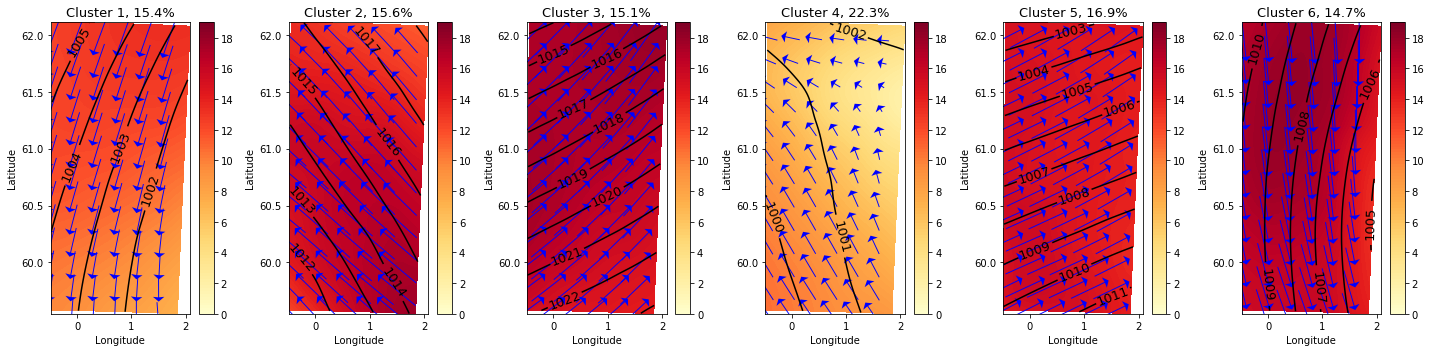}
    \caption{Clustering on wind velocity data in the smaller Shetland Islands domain}
\end{subfigure}
\begin{subfigure}{0.95\textwidth}
    \centering
    \includegraphics[width = 0.8\textwidth]{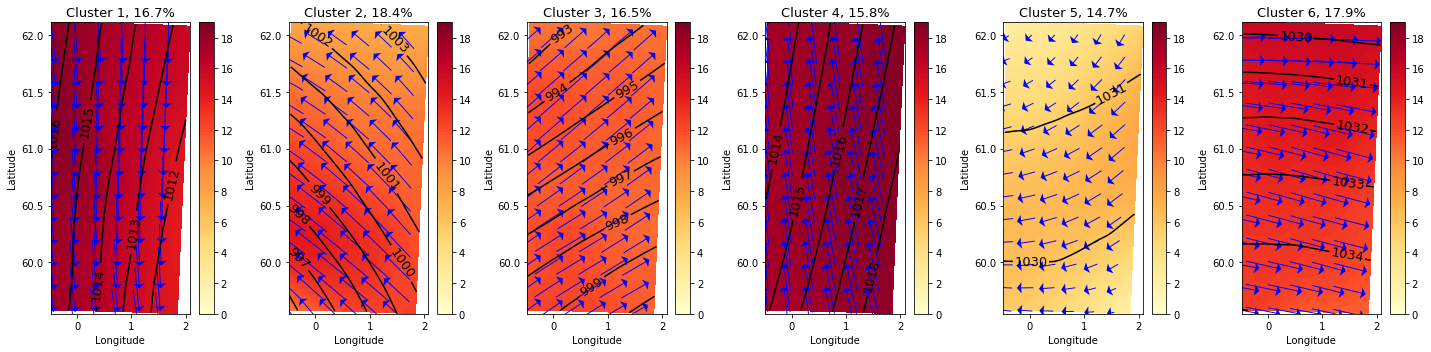}
    \caption{Average pressure and wind velocity for the patterns found by clustering on wind velocity data in the Europe-wide domain.}
\end{subfigure}
\begin{subfigure}{0.95\textwidth}
    \centering
    \includegraphics[width = 0.8\textwidth]{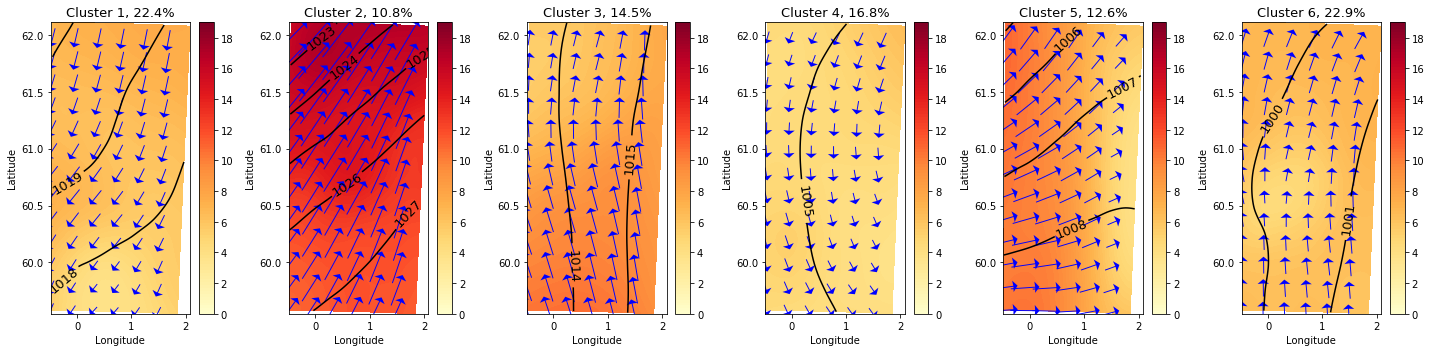}
    \caption{Average pressure and wind velocity for the patterns found by clustering on sea-level pressure data in the Europe-wide domain.}
\end{subfigure}
\caption{Shetland Island domain cluster and averages using Europe-wide domains. Filled contours represent the wind speed, black contours the pressure and the arrows the wind direction with size reflecting the wind speed magnitude.}\label{fig:cluster_d02_shet}
\end{figure}

Figure \ref{fig:cluster_d02_shet} shows the same analysis but for the small Shetland Islands domain with (a) showing the clusters when clustering on wind velocity data in that domain, and (b) the clusters when clustering on wind velocity and (c) on pressure in the Europe-wide domain. In this domain, the cluster sets are more similar than in the Denmark domain, with the two cluster sets found by clustering on wind velocity again more similar especially in magnitude, than that found by clustering on pressure. There are however still stark differences in wind patterns, with for example (a) having two clusters with north-westerly wind and two with north-easterly wind; (b) having one cluster each for north-westerly and north-easterly wind and (c) having only one cluster for north-westerly wind and no cluster for north-easterly wind. The variety of weather patterns shown in Figures \ref{fig:cluster_d01}, \ref{fig:cluster_d02_den} and \ref{fig:cluster_d02_shet} thus justify our decision to consider multiple variables and domains to predict wind energy and downstream wakes rather than to follow the literature and focus only on pressure on a Europe-wide domain. In Section \ref{sec:patterns}, we show how these different weather patterns lead to power and downstream wake estimates which have different levels of accuracy, with weather patterns found by clustering on wind velocity being more accurate than those found by clustering on pressure.

\begin{figure}[!t]
    \centering
        \includegraphics[width = 0.45\textwidth]{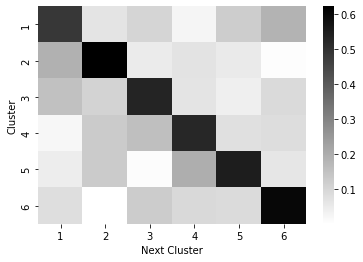}
    \includegraphics[width = 0.45\textwidth]{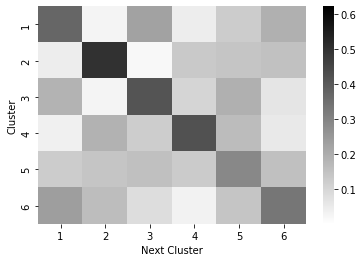}
    \caption{Comparison of probability transition matrices for clusters from wind velocity on Europe-wide domain (left) and smaller Denmark domain (right). The probability of staying the same pattern for consecutive days is greater for the larger domain than the smaller domain.}
    \label{fig:prob_transition}
\end{figure}

Finally, an important factor to consider when using clustering for weather patterns is how long a particular cluster persists. If a weather pattern persists over a long period of time then one could argue that it is more than a pattern and actually a weather regime. Figure \ref{fig:prob_transition} shows the probability transition matrix for clusters from wind velocity for the Europe-wide and smaller Denmark domain. The diagonal line from the top-left to the bottom-right of the matrix is the probability of staying in the same weather cluster for consecutive days (\textit{i.e.} no transition). The probability of no transition is clearly higher for the Europe-wide domain than for the smaller Denmark domain. This means the weather patterns identified by clustering on the Europe-wide domain persist for longer. For brevity, we have not included the probability transition matrices for the other cluster sets here but they show a similar result. This is unsurprising because it is only at a continental scale that it is possible to identify persistent weather patterns; whereas on a local scale weather patterns are more variable due to local changes and are therefore less likely to persist. Note that the latter does not necessarily make the clusters on the smaller domain less valid but means we can not refer to the patterns found by k-means clustering as regimes.

\section{Using weather patterns to determine power output and downstream wakes}\label{sec:patterns}
The weather patterns found in the previous section are already of interest from a meteorological perspective and in this section we show how they can also be of great use and interest for wind energy forecasting. To accurately verify our long-term power and wake predictions, we would need to compare the clustering results to the results from running WRF for every date-time in the dataset. However, this would be very computationally expensive and therefore we only run WRF for every datapoint in one year. Specifically, we choose to focus on 2007 because this was a year where there was large investment in wind energy \citep{council2007gwec}. We thus also use our clustering approach to predict total power output and downstream wakes in 2007, emphasising that the clusters are still those found in Section \ref{sec:clustering} over the entire 2000--2009 dataset and therefore are not necessarily datapoints from 2007.

\subsection{Power production predictions}
\label{sec:power_production_predictions}
\begin{figure}[!t]
    \centering
    \includegraphics[width=0.4\textwidth]{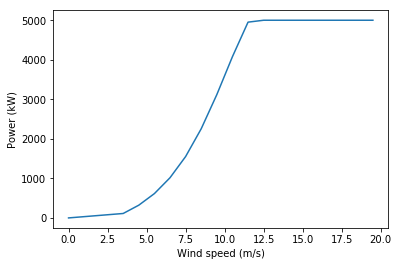}
    \caption{Power curve for wind turbines used in hypothetical wind farm.}
    \label{fig:power_curve}
\end{figure}

\begin{figure}[!t]
\centering
    \begin{subfigure}{0.48\textwidth}
    \includegraphics[width= \textwidth]{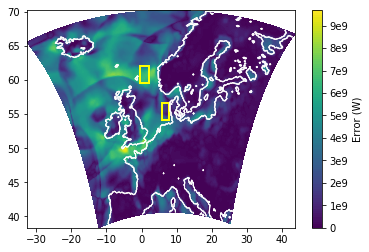}
    \caption{Clustering on wind velocity data from 2000--2009 (Europe).}
    \label{fig:err_wind_d01_power_curve}
    \end{subfigure}
    \begin{subfigure}{0.48\textwidth}
    \includegraphics[width = \textwidth]{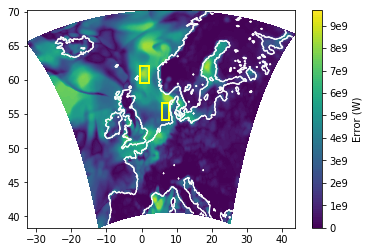}
    \caption{Clustering on pressure data from 2000--2009 (Europe).}
    \label{fig:err_press_d01_power_curve}
    \end{subfigure}
    \caption{Difference between the total power forecast when applying a power curve to the real data across the Europe-wide domain and a simple weighted sum of the power curve applied to the cluster centres. The small domains are marked as yellow boxes on the figures.}\label{fig:power_curve_err}
\end{figure}

Before running any WRF simulations, we first investigate the wind energy climate of the Shetland Islands domain and Denmark domain, using the simple empirical approach of a power curve. Every wind turbine has an associated power curve, which is the expected relationship between wind speed and power. Figure \ref{fig:power_curve} shows the power curve for the wind turbines used in our hypothetical wind farm. We can apply this power curve to both the full set of wind speeds from the reanalysis data, and the cluster centres. Figure \ref{fig:power_curve_err} shows the difference between the total power predicted when applying the power curve to the full reanalysis data, and that predicted when applying the power curve to the Europe-wide cluster centres and then using the simple weighted sum (\ref{eq:simple}). In general, the difference from clustering on pressure data (MAE: \SI{1.9e9}{W}) is larger than that from clustering on wind velocity data (MAE: \SI{1.75e9}{W}). Of particular note though is that in both (a) and (b), the difference in the Shetland Islands domain is greater than in the Denmark domain (these domains are marked by yellow boxes on the figure). This suggests that the wind energy climate is more varied in the Shetland domain and therefore that it may be more difficult to obtain good predictions for wind power than in the Denmark domain.

\begin{figure}[!t]
    \centering
    \begin{subfigure}{0.48\textwidth}
    \includegraphics[width= \textwidth]{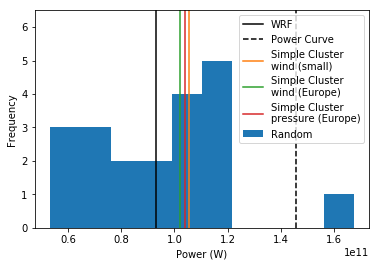}
    \caption{Denmark simple (\ref{eq:simple}).}
    \label{fig:clust_power_dist_den_simple}
    \end{subfigure}
    \begin{subfigure}{0.48\textwidth}
    \includegraphics[width = \textwidth]{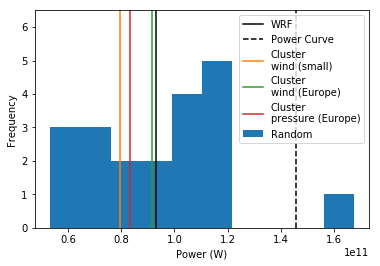}
    \caption{Denmark complex (\ref{eq:complex wake}).}
    \label{fig:clust_diff_u_dist_den_complex}
    \end{subfigure}
        \begin{subfigure}{0.48\textwidth}
    \includegraphics[width= \textwidth]{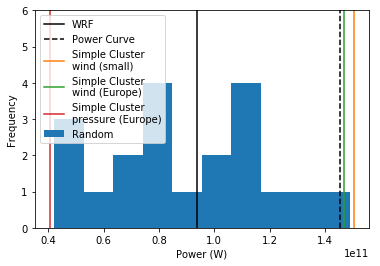}
    \caption{Shetland simple (\ref{eq:simple}).}
    \label{fig:clust_power_dist_shet_simple}
    \end{subfigure}
    \begin{subfigure}{0.48\textwidth}
    \includegraphics[width = \textwidth]{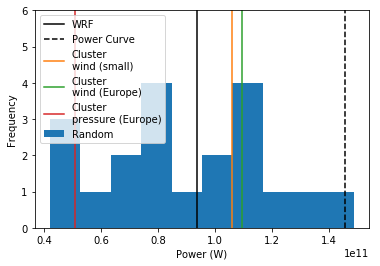}
    \caption{Shetland Islands complex (\ref{eq:complex wake}).}
    \label{fig:clust_power_dist_shet_complex}
    \end{subfigure}
    \caption{Comparing the correct power output from WRF with estimates when using the simple (\ref{eq:simple}) and complex (\ref{eq:complex wake}) weighted sums for each cluster and a distribution of values when randomly selecting 6 datapoints from the dataset 365 times.}\label{fig:power_comparison}
\end{figure}

This simple power curve approach does not provide any information on downstream wakes and the power predictions are also not as accurate as those obtained with WRF. We therefore seek more accurate predictions by running WRF for the closest datapoint to the cluster centres, as detailed in Section \ref{sec:methodology}. Figure \ref{fig:power_comparison} compares the total power prediction for Denmark and the Shetland Islands from using the simple weighted sum (\ref{eq:simple}) and more complex weighted sum (\ref{eq:complex wake}) of the WRF cluster runs, with the prediction from running WRF on each day. The total power predictions using the complex weighted sums are closer to the real WRF value than those using the simple weighted sums, particularly for the Shetland Islands, justifying our more complex post-processing technique. For the Denmark farm, the best power prediction is obtained by the clusters found by clustering on wind velocity in the Europe-wide domain, whereas the best power prediction for the Shetland Islands farm is obtained by clustering on the wind speed in the smaller domain. This is consistent with the findings from Figure \ref{fig:power_curve_err}, which suggested that the clusters found in the larger domain would perform worse at the Shetland farm than the Denmark farm. This is particularly true when clustering on pressure and highlights the fact that pressure is not the most appropriate variable to cluster on for the purposes of wind power prediction. However, when clustering on wind velocity, even for a case which appeared extreme in Figure \ref{fig:power_curve_err}, the difference in the power prediction is relatively small between clustering on the small and large domain.

To highlight the need for using both WRF and clustering to obtain good predictions, Figure \ref{fig:power_comparison} includes two other benchmarks: (i) the distribution of the total power prediction if six datapoints are randomly selected from the dataset 365 times, and (ii) applying the power curve in Figure \ref{fig:power_curve} to the reanalysis wind speeds. Note that for the first benchmark, we choose to sample 365 times to maximise the probability that each datapoint will be selected at least once. The wide distribution of values produced as a result, in all cases in Figure \ref{fig:power_comparison}, emphasises the benefit of the clustering approach for selecting six representative datapoints and aggregating the results from their corresponding WRF simulations, compared with taking a random sample of six datapoints. For the second benchmark, our WRF clustering results give a more accurate power prediction than applying the power curve directly to the reanalysis data. The exception is for the Shetland Islands using the simple weighted sum where all prediction methods produce similarly poor estimates of the total power. We attribute this to the complex wind energy climate in the Shetland Islands region. Note however that our complex post-processing approach still performs well for the Shetland Islands case study. This therefore shows the importance of using the more complex post-processing approach, as well as using the WRF model, to obtain good predictions of power.

To summarise, this subsection shows that combining our WRF clustering approach with the more complex weighted sum post-processing technique results in accurate predictions of power output. Notably, these predictions are accurate even for clusters found from clustering wind velocity on the Europe-wide domain. Hence, by performing clustering analysis over the full European domain only once, it is possible to find wind farm power output estimates for wind farm sites across Europe, which are at least as accurate as the estimates obtained by clustering on wind velocity over the small domains. This generalisability provides considerable time-saving benefits, making our method very efficient and easily applicable to novel situations.

\subsection{Downstream wake predictions}

\begin{figure}[!t]
\centering
\begin{subfigure}{0.9\textwidth}
    \centering
    \includegraphics[width= \textwidth]{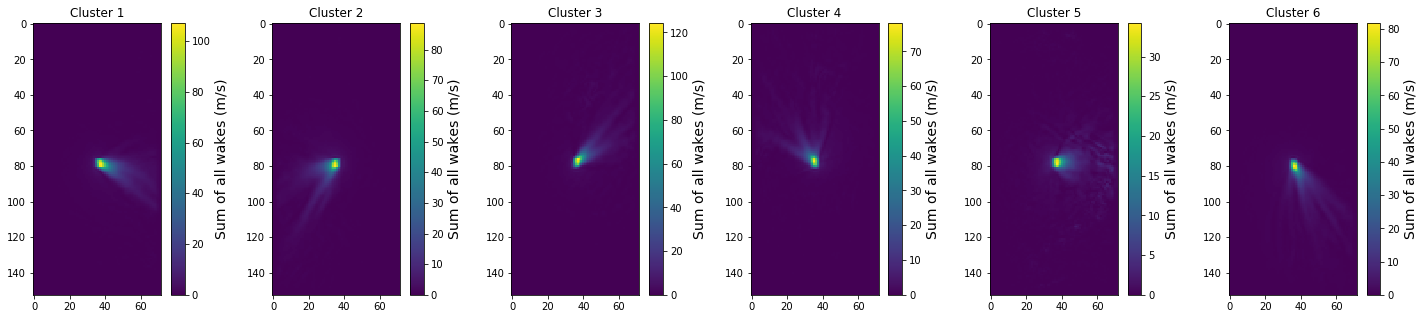}
    \caption{Real wake prediction generated using WRF for each of the clusters.}
\end{subfigure}
\begin{subfigure}{0.9\textwidth}
\centering
    \includegraphics[width= \textwidth]{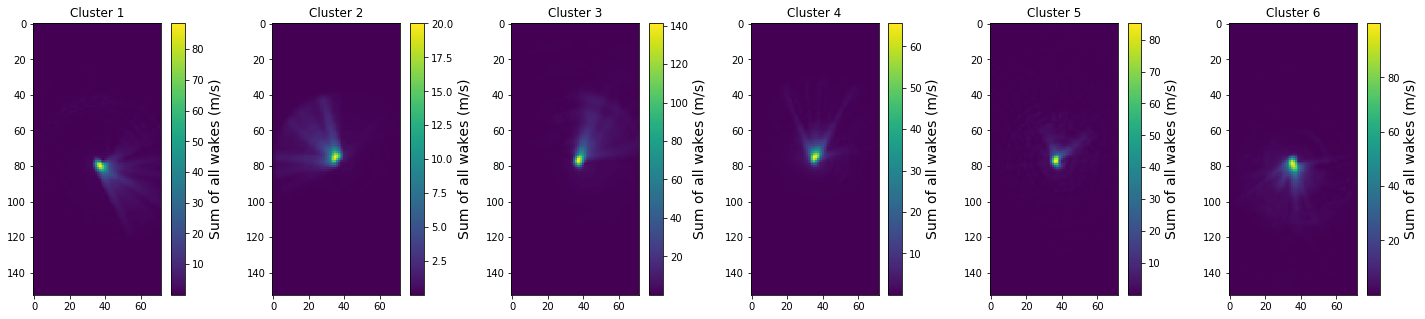}
    \caption{Wake prediction using complex weighted sum \eqref{eq:complex wake} of cluster WRF results for each of the clusters.}
    \end{subfigure}
        \begin{subfigure}{0.9\textwidth}
    \centering
    \includegraphics[width= \textwidth]{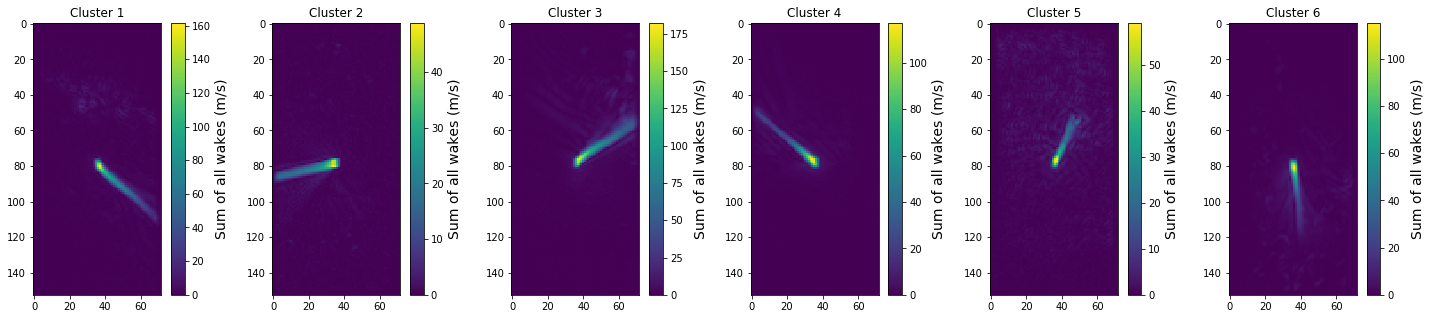}
    \caption{Wake prediction using simple weighted sum \eqref{eq:simple} of cluster WRF results for each of the clusters.}
    \end{subfigure}
    \caption{Wake predictions in the small Denmark domain using clusters found by clustering on wind velocity in the small Denmark domain. Here we show the total sum of the wake field magnitude for all datapoints in the cluster.}
    \label{fig:denmark_wind_d02_diffu}
\end{figure}

\begin{figure}[!t]
\centering
\begin{subfigure}{0.9\textwidth}
    \centering
    \includegraphics[width= \textwidth]{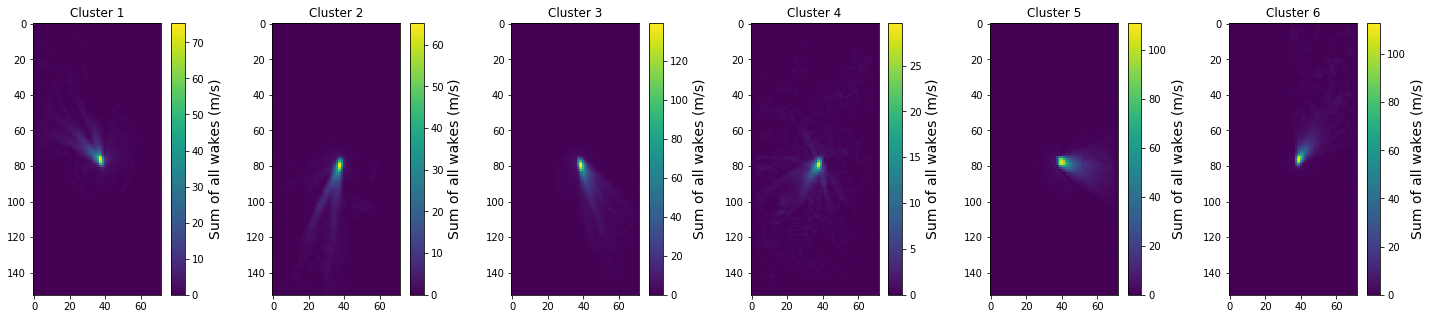}
    \caption{Real wake prediction using WRF for each of the clusters.}
\end{subfigure}
\begin{subfigure}{0.9\textwidth}
\centering
    \includegraphics[width= \textwidth]{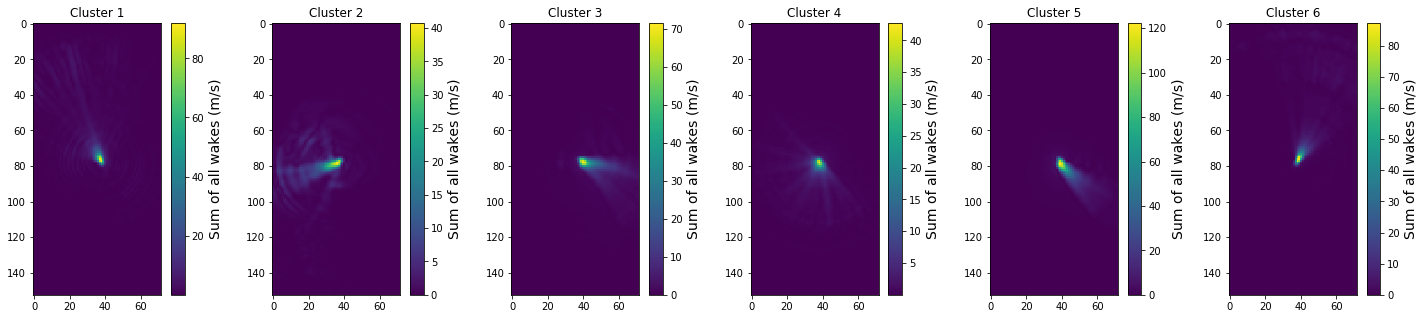}
    \caption{Wake prediction using complex weighted sum \eqref{eq:complex wake} of cluster WRF results for each of the clusters.}
    \end{subfigure}
    \begin{subfigure}{0.9\textwidth}
    \centering
    \includegraphics[width= \textwidth]{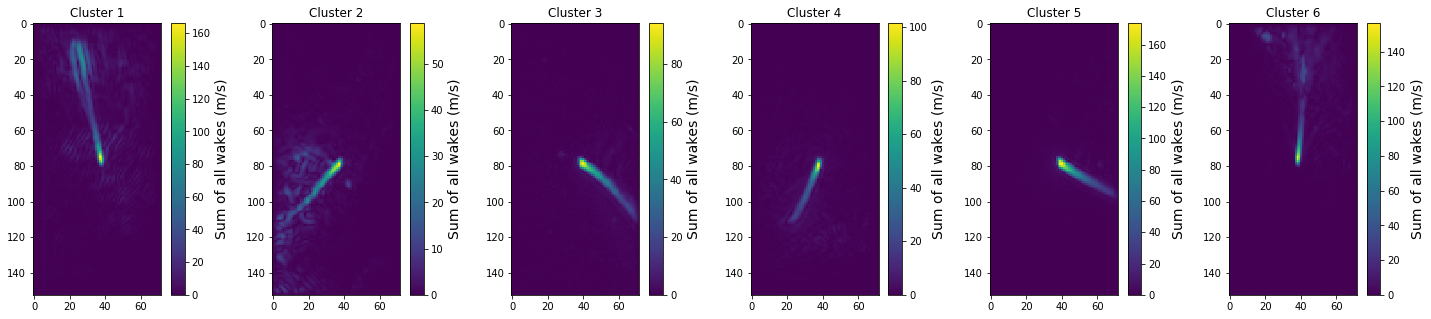}
    \caption{Wake prediction using simple weighted sum \eqref{eq:simple} of cluster WRF results for each of the clusters.}
    \end{subfigure}
    \caption{Wake predictions in the small Shetland Islands domain using clusters found by clustering on wind velocity in the Shetland Islands domain. Here we show the total sum of the wake field magnitude for all datapoints in the cluster.}
    \label{fig:shet_wind_d02_diffu}
\end{figure}

\begin{figure}[!t]
\centering
\begin{subfigure}{0.9\textwidth}
    \centering
    \includegraphics[width= \textwidth]{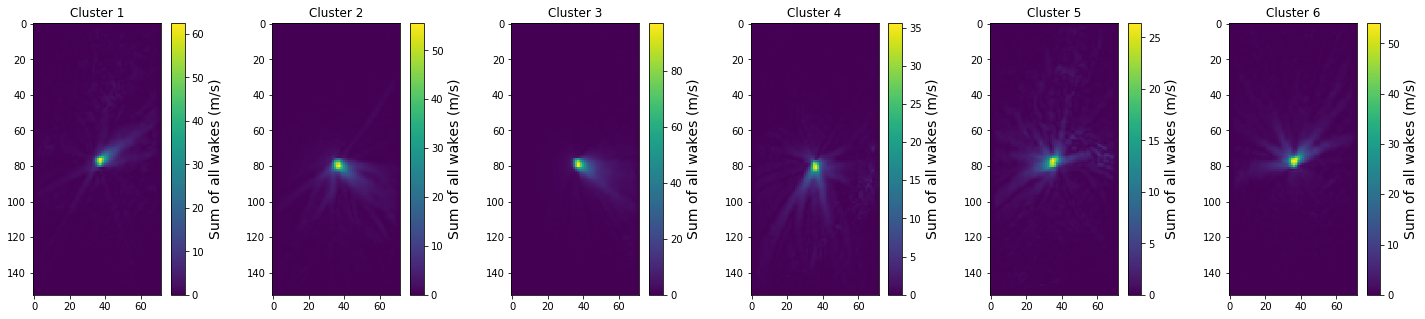}
    \caption{Real wake prediction generated using WRF for each of the clusters.}
\end{subfigure}
\begin{subfigure}{0.9\textwidth}
\centering
    \includegraphics[width= \textwidth]{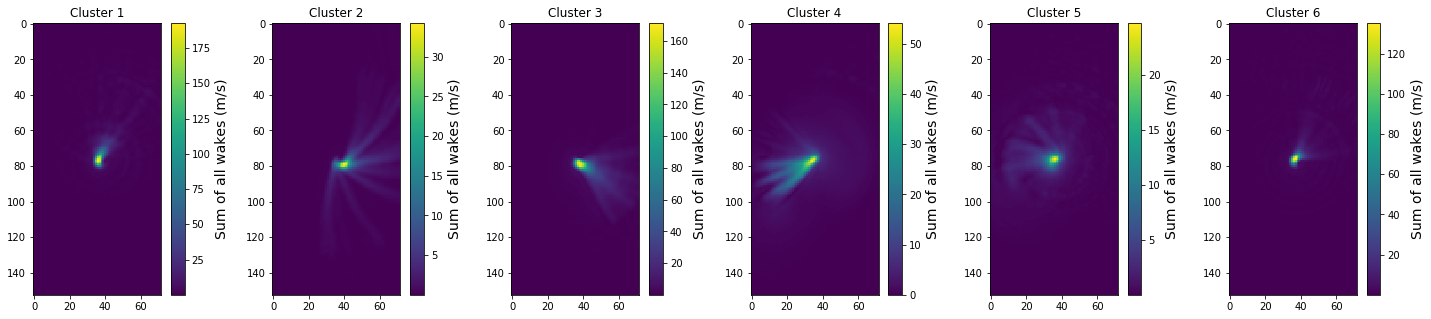}
    \caption{Wake prediction using complex weighting \eqref{eq:complex wake} of cluster WRF results for each of the clusters.}
    \end{subfigure}
    \caption{Wake predictions in the small Denmark domain using clusters found by clustering on wind velocity in the Europe-wide domain. Here we show the total sum of the wake field magnitude for all datapoints in the cluster.}
    \label{fig:denmark_wind_d01_diffu}
\end{figure}

\begin{figure}[!t]
\centering
\begin{subfigure}{0.9\textwidth}
    \centering
    \includegraphics[width= \textwidth]{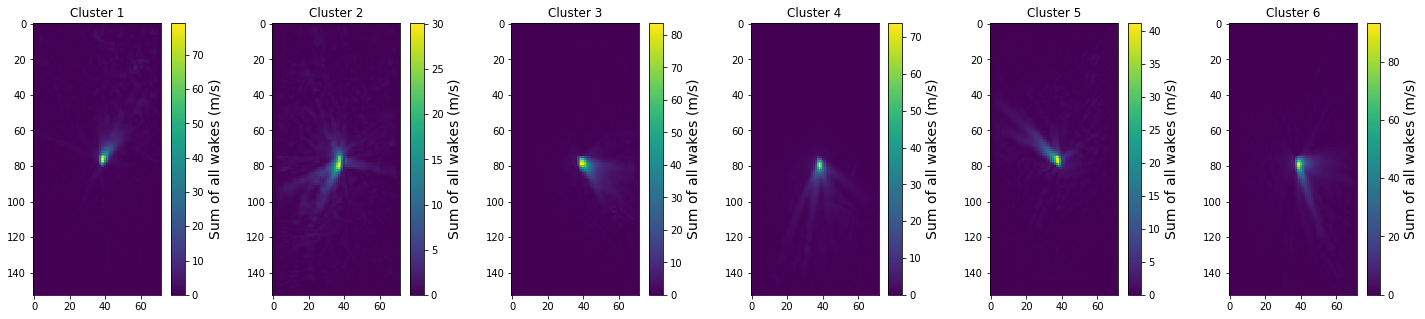}
    \caption{Real wake prediction using WRF.}
\end{subfigure}
\begin{subfigure}{0.9\textwidth}
\centering
    \includegraphics[width= \textwidth]{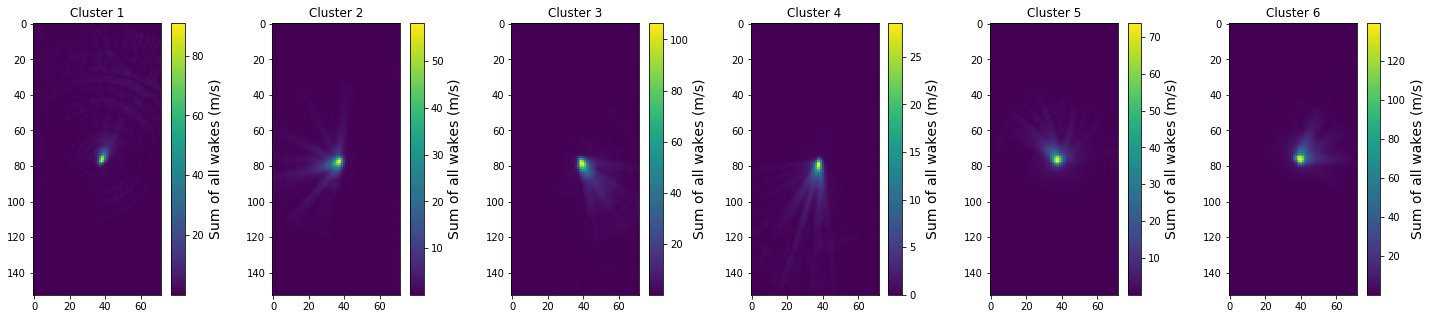}
    \caption{Wake prediction using complex weighted sum \eqref{eq:complex wake} of cluster WRF results}
    \end{subfigure}
    \caption{Wake predictions in the small Shetland Islands domain using clusters found by clustering on wind velocity in the Europe-wide domain. Here we show the total sum of the wake field magnitude for all datapoints in the cluster.}
    \label{fig:shet_wind_d01_diffu}
\end{figure}

\begin{figure}[!t]
\centering
\begin{subfigure}{0.9\textwidth}
    \centering
    \includegraphics[width= \textwidth]{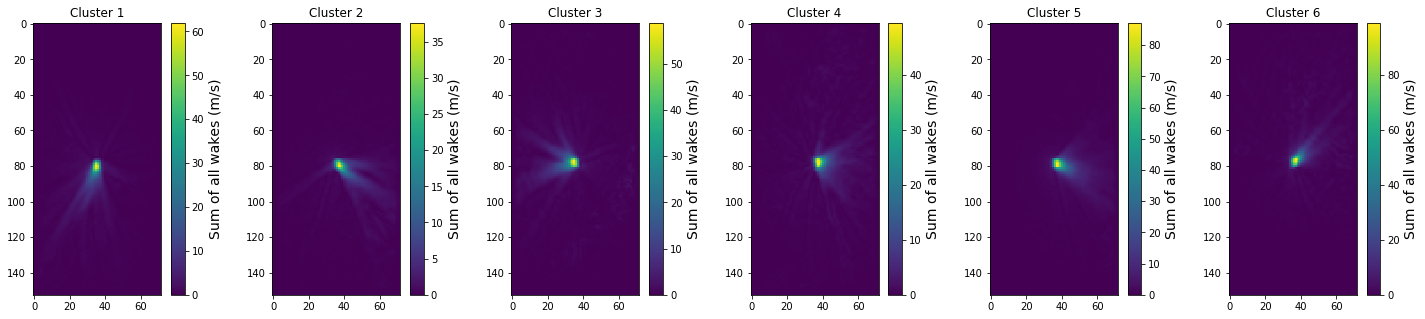}
    \caption{Real wake prediction using WRF for each of the clusters.}
\end{subfigure}
\begin{subfigure}{0.9\textwidth}
\centering
    \includegraphics[width= \textwidth]{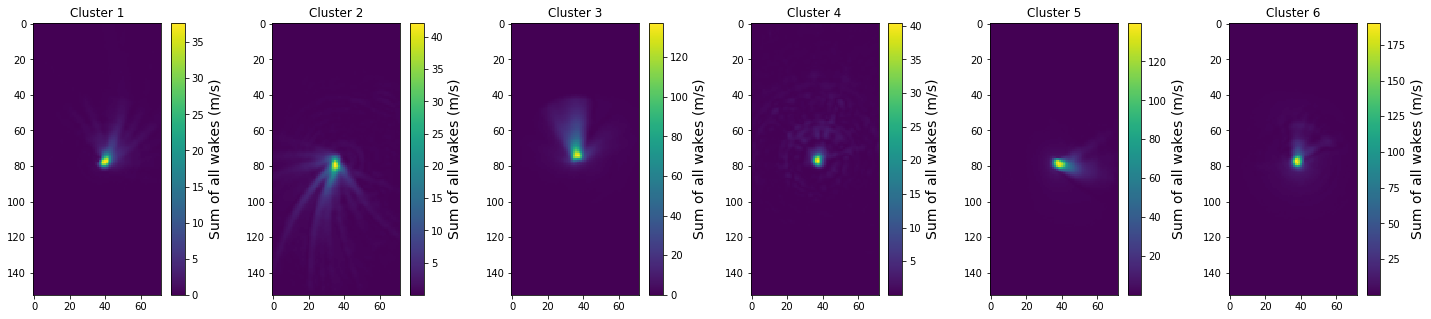}
    \caption{Wake prediction using complex weighted sum \eqref{eq:complex wake} of cluster WRF results for each of the clusters.}
    \end{subfigure}
    \caption{Wake predictions in the small Denmark domain using clusters found by clustering on pressure in the Europe-wide domain. Here we show the total sum of the wake field magnitude for all datapoints in the cluster.}
    \label{fig:denmark_press_d01_diffu}
\end{figure}

\begin{figure}[!t]
\centering
\begin{subfigure}{0.9\textwidth}
    \centering
    \includegraphics[width= \textwidth]{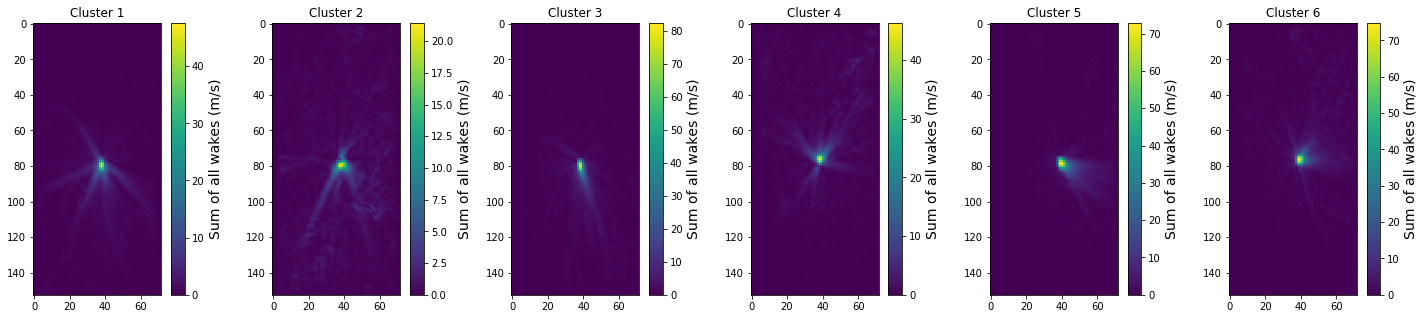}
    \caption{Real wake prediction using WRF for each of the clusters.}
\end{subfigure}
\begin{subfigure}{0.9\textwidth}
\centering
    \includegraphics[width= \textwidth]{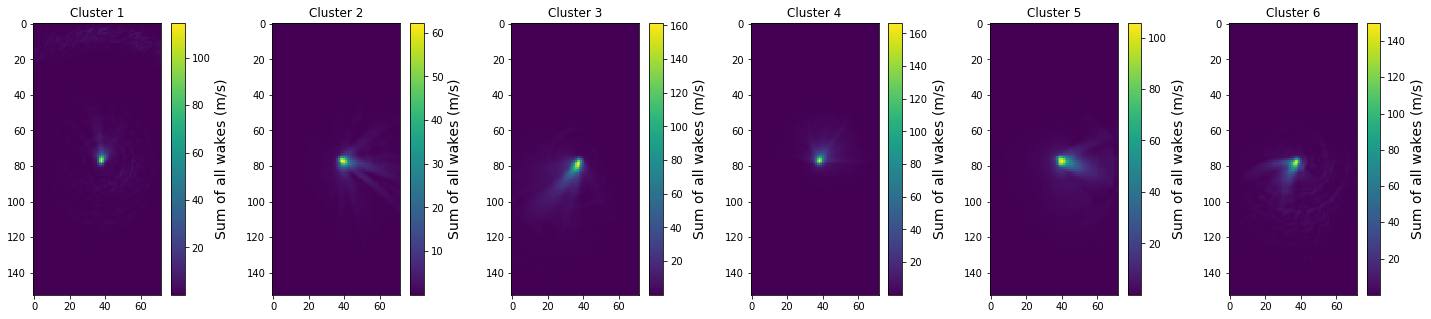}
    \caption{Wake prediction using complex weighted sum \eqref{eq:complex wake} of cluster WRF results for each of the clusters.}
    \end{subfigure}
    \caption{Wake predictions in the small Shetland Islands domain using clusters found by clustering on pressure in the Europe-wide domain. Here we show the total sum of the wake field magnitude for all datapoints in the cluster.}
    \label{fig:shet_press_d01_diffu}
\end{figure}

Crucial to quantifying wind loss effects due to wakes from nearby farms is the estimation of the mean downstream wake from a given wind farm. However, as discussed in Section \ref{sec:methodology}, aggregating downstream wakes is more complicated than combining power estimates because there is a spatial component as well as a temporal component. For the Denmark domain, the mean (over the year 2007) WRF-derived wake predictions are compared with the combined wake predictions obtained using the clustering approach in Figures \ref{fig:denmark_wind_d02_diffu}, \ref{fig:denmark_wind_d01_diffu} and \ref{fig:denmark_press_d01_diffu} for clustering on wind velocity in the small domain, wind velocity on the Europe-wide domain and pressure on the Europe-wide domain, respectively. Figures \ref{fig:shet_wind_d02_diffu}, \ref{fig:shet_wind_d01_diffu} and \ref{fig:shet_press_d01_diffu} show the same but for the Shetland Islands domain. In order to be able to easily compare wake patterns predicted by the different approaches, we show the wake predictions for each cluster separately, i.e. the mean wakes over days assigned to each cluster. Note that the downside of displaying the results like this is that because the clusters are different depending on the variables clustered on (see Section \ref{sec:clustering}), it is incorrect to compare the wake patterns in Figures \ref{fig:denmark_wind_d02_diffu} with those in Figure \ref{fig:denmark_wind_d01_diffu} or Figure \ref{fig:denmark_press_d01_diffu} etc. Note further that for reasons of brevity, we only include the predictions from the simple weighted sum approach (\ref{eq:simple}) in Figures \ref{fig:denmark_wind_d02_diffu} and \ref{fig:shet_wind_d02_diffu}, which are the result of clustering on wind velocity on the small domain for Denmark and the Shetland Islands respectively. Both figures clearly show that the shape, direction and magnitude of the wake prediction is more accurate when using the complex weighted sum (\ref{eq:complex wake}) compared to the simple weighted sum (\ref{eq:simple}), and in general the complex weighted sum prediction closely resembles the real wake prediction from WRF. This is particularly evident for Cluster 3 in Figure \ref{fig:denmark_wind_d02_diffu} and Cluster 6 in Figure \ref{fig:shet_wind_d02_diffu}, where the direction and shape of the wake is completely wrong using the simple weighted sum, but the prediction for the complex weighted sum is almost indistinguishable in shape and direction from the real wake prediction from WRF.

The results of section \ref{sec:power_production_predictions} show that clustering on wind velocity in the Europe-wide domain produces good predictions of long-term power output. Figures \ref{fig:denmark_wind_d01_diffu} and \ref{fig:shet_wind_d01_diffu} show that this clustering method can also result in reasonable wake predictions. For the Shetland Islands domain, the predictions are of a similar order of accuracy as those in Figure \ref{fig:shet_wind_d02_diffu} (both sets of predictions have an approximate average correlation of 0.6 with the real clusters). The exception in Figure \ref{fig:shet_wind_d01_diffu} is Cluster 6, where the direction and magnitude of the wake field are incorrect. However, for the Denmark domain, most of the cluster predictions are not as accurate as in Figure \ref{fig:denmark_wind_d02_diffu} (see for example Clusters 1 and 3 in Figure \ref{fig:denmark_wind_d01_diffu} where the magnitudes of the wakes are substantially different). 

Finally, Figures \ref{fig:denmark_press_d01_diffu} and \ref{fig:shet_press_d01_diffu} show that clustering on pressure on the Europe-wide domain results in poor predictions of downstream wake in both the Denmark and Shetland Islands domains. The magnitude and direction of the wake is incorrect for almost all clusters. For example in Cluster 3 in Figure \ref{fig:denmark_press_d01_diffu}, the wake direction should be to the west but is to the north in the WRF clustering-approximation and the magnitude of the approximation is more than twice as large as that of the real value. 

In summary, the analysis in this subsection has shown that accurate predictions of downstream wakes can be obtained by clustering on wind velocity, but clustering on pressure results in incorrect predictions. This is significant because previous works on weather patterns have focussed on clustering on pressure rather than on wind velocity \citep[e.g.][]{neal2016flexible,cheneka2020impact,garrido2020impact}. Furthermore, in contrast to the power predictions of section \ref{sec:power_production_predictions}, the downstream wake predictions are much more accurate in magnitude, direction and shape when clustering on variables from the small local-scale domain, compared to from the Europe-wide continent-scale domain. This is to be expected because the direction and shape of the downstream wake is particularly sensitive to weather systems on a local scale.

\subsection{Effect of the wind farms on the wind pattern clusters}
\begin{figure}[!t]
    \centering
    \includegraphics[width = 0.9\textwidth]{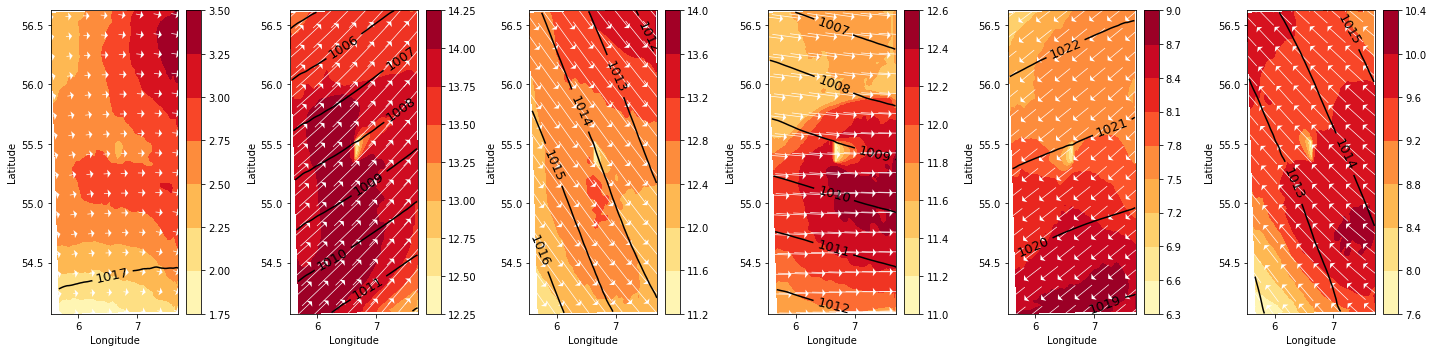}
    \caption{Clustering on wind velocity in the small Denmark domain at 15:00 every day in 2007 with a wind farm included.}
    \label{fig:cluster_with_wf}
    \includegraphics[width = 0.9\textwidth]{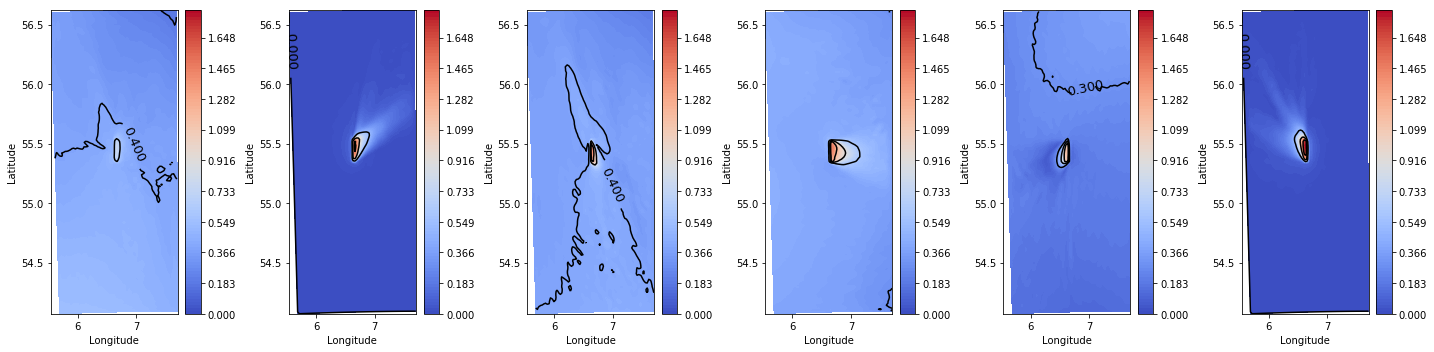}
    \caption{Difference in wind velocity magnitude when clustering with and without a wind farm.}
    \label{fig:cluster_diff_wf}    
\end{figure}

As demonstrated above, the presence of wind farms leads to downstream wakes. This in turn alters the wind climate. Hence, if the clustering analysis on the small domains is repeated on wind fields in the presence of wind farms, the clusters identified in Section \ref{sec:clustering} will change and datapoints may become assigned to different clusters. Our original clustering is performed on reanalysis data spanning 10 years. However, repeating the analysis in the presence of wind farms requires a set of WRF model runs with farms included. For reasons of computational cost, we do not run WRF for the full 10-year period. We instead re-use the set of WRF runs spanning the year 2007, which were introduced above for the purposes of validating power and wake estimates. In order to avoid inconsistencies caused by comparing clustering results based on reanalysis and WRF-derived datasets, we also repeat the clustering analysis using WRF outputs from 2007 without a farm, and make comparisons between the clusters from these two WRF-derived datasets. Note that this decision is justified because the clusters derived from the WRF-derived datasets are different to the original ones: for example, Figure \ref{fig:cluster_with_wf} shows two westerlies when clustering on the WRF-derived dataset and Figure \ref{fig:cluster_d02_den}a shows no westerlies when clustering in the original ERA5 dataset.

Figure \ref{fig:cluster_with_wf} shows the cluster centres obtained when clustering on these wind velocity WRF outputs in the small Denmark domain in the presence of a wind farm. In all six clusters in the figure, the wind farm is easily detectable, but for ease, we have also included Figure \ref{fig:cluster_diff_wf}, which shows the difference between the cluster centres with and without a wind farm. After just three hours of simulation, the presence of the wind farm clearly results in a reduction in velocity compared to the surrounding area, though the general cluster patterns remain the same. Even so, this is sufficient for nine datapoints (out of 365) to change cluster label due to the presence of the wind farm. This indicates that the impact of other wind farms may mean that any local-scale clustering approach for the prediction of wakes must be repeated for each new wind farm added following a roll-out type sequence. Note that there is no need for repeated clustering for the power predictions as these are made using the Europe-wide domain cluster, which are unaffected.

\section{Conclusion}\label{sec:conclusion}
This work shows that accurate long-term predictions of downstream wakes and wind power production can be obtained by only running simulations for key weather patterns identified using unsupervised clustering methods applied to wind velocity data. Consistent with previous work, we found that six clusters are sufficient to identify the key patterns in weather data, which means the numerical model (in our case WRF) only needs to be run six times for each cluster-based prediction. This results in a fast, accurate and effective approach for long-term predictions.

The most accurate cluster-based predictions for downstream wake and power are obtained when clustering on wind velocity data. This is a significant result because previous literature has focused on weather patterns identified by clustering on surface pressure. Moreover, for long-term power production predictions, it is sufficient to use the clusters found by clustering for wind velocity in the Europe-wide domain. Thus it is only necessary to perform the clustering analysis once for all wind farms in Europe. For the downstream wakes, however, it is necessary to cluster on the local-scale domains. This is to be expected due to the local scale nature of the phenomenon. Furthermore, all cluster-based predictions are significantly improved by our novel post-processing approach, which makes full use of the information in the underlying reanalysis data. 

The ultimate goal of the approach outlined in this work is to contribute to a suite of design tools, which will enable accurate long-term prediction of the power production and downstream wakes from future wind farms. In particular, these tools will be able to accurately predict power losses incurred when wind farms are installed in close proximity. We have shown in this work that the downstream wakes affect the local-scale clusters themselves. Thus in further work we will explore how the addition of large numbers of new wind farms will further affect the cluster patterns, and hence the long-term predictions of downstream wakes and power loss. This will also help to determine the impact of wind farms on local weather patterns, which is a key consideration for the development of wind energy. 

\section*{Data and Code Availability}
The relevant code for the clustering framework presented in this work is stored at \url{https://github.com/mc4117/Clustering_wind_resources.git}. The ERA5 reanalysis data used to determine the weather patterns was downloaded from the Copernicus Climate Data Store \url{https://cds.climate.copernicus.eu/}. The WRF model used for the numerical simulations can be found at \url{https://www.mmm.ucar.edu/models/wrf}.

\section*{Acknowledgements}
The authors would like to acknowledge funding from Shell Research Ltd, UK, and the UK Engineering and Physical Sciences Research Council (project EP/R029423/1). The authors would also like to thank the Imperial College London Research Computing Service for the facilities they provide which were used to conduct this research.

\bibliographystyle{cas-model2-names}
\typeout{}
\bibliography{references}

\end{document}